\documentclass{article}

\usepackage{iclr2026_conference,times}
\usepackage{amsmath}
\usepackage{amssymb}
\usepackage{mathtools}
\usepackage{bm}
\usepackage{pifont}

\usepackage{hyperref}
\usepackage{url}
\usepackage{cleveref}

\usepackage{graphicx}
\usepackage{subcaption}
\usepackage{booktabs}
\usepackage{multirow}
\usepackage{wrapfig}
\usepackage{float}
\usepackage{colortbl}
\usepackage{xcolor}
\usepackage{caption}
\usepackage{tikz}

\usepackage{soul}        % 高亮与删除线
\usepackage{acronym}     % 缩写管理
\usepackage{fancybox}    % 边框盒子
\usepackage{epigraph}    % 题注/格言
\usepackage{dirtytalk}   % 引号转换
\usepackage{fontawesome} % 图标字体
\usepackage{enumitem}    % 列表控制
\usepackage{comment}     % 大段注释

\definecolor{best}{HTML}{C8E6C9}
\definecolor{best2}{HTML}{BBDEFB}

\title{Large Depth Completion Model from Sparse Observations}

\author{Zhu Yu$^{1}$\thanks{Internship at Tongyi Lab} \quad
	Zhengyi Zhao$^{2}$ \quad
	Runmin Zhang$^{1}$ \quad
	Lingteng Qiu$^{2}$ \quad
	Kejie Qiu$^{2}$ \quad \\ \bf
	Yisheng He$^{2}$  \quad
	Siyu Zhu$^{3}$ \quad
	Zilong Dong$^{2\dagger}$\quad
	Si-Yuan Cao$^{4,5,6}$\thanks{Corresponding author}\quad
	Hui-Liang Shen$^{1}$\quad
	\\[2pt]
	\small{$^1$Zhejiang University}\hspace{1em}
	\small{$^2$Tongyi Lab, Alibaba Group}\hspace{1em}
	\small{$^3$Fudan University}
	\\
	\small{$^4$Ningbo Innovation Center, Zhejiang University}\hspace{1em} 
	\small{$^5$NingboTech University}
	\\
	\small{$^6$Jinhua Institute of Zhejiang University}
	\\
	{\tt \small{\{yu\_zhu, cao\_siyuan\}@zju.edu.cn}}
	\\
	{\tt\small
		\faGithubAlt~\textbf{Project Page:} \href{https://pkqbajng.github.io/ldcm/}{\texttt{https://pkqbajng.github.io/ldcm/}}
	}
}

\hypersetup{
	colorlinks=true,  % 开启文字着色
	linkcolor=blue,  % 
	citecolor=green,   % 
	urlcolor=red     % 网址链接颜色
}

\newcommand{\eg}{e.g.}

\iclrfinalcopy
\begin{document}
\maketitle
\thispagestyle{fancy}
\fancyhead[L]{Published as a conference paper at ICLR 2026} 
\ificlrfinal\else\lhead{Under review as a conference paper at ICLR 2026}\fi
\vspace{-6mm}
\begin{figure}[h]
	\centering
	\includegraphics[width=\linewidth]{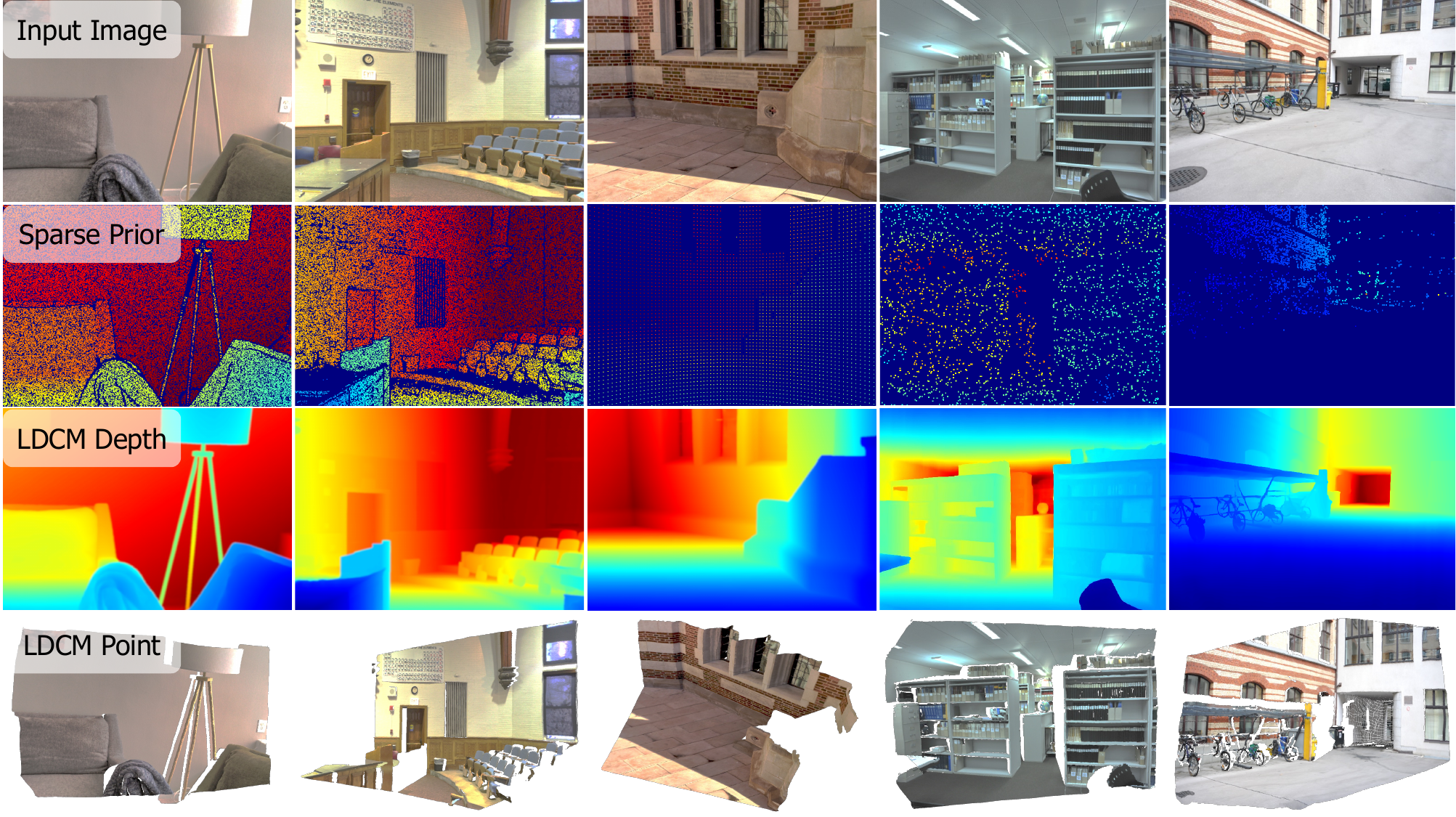}
	\begin{subfigure}[t]{0.4\linewidth}
		\centering
		\includegraphics[width=\linewidth]{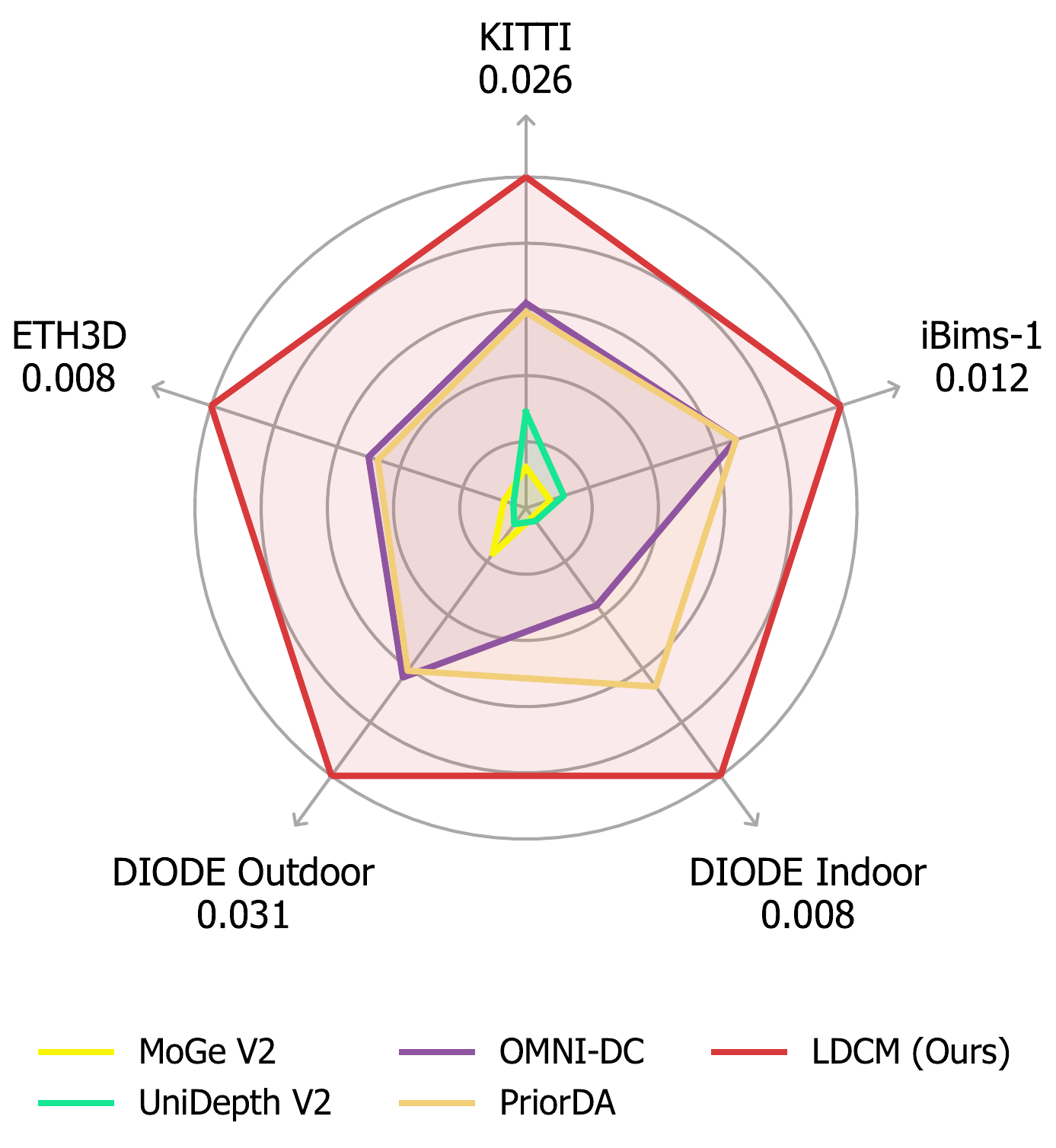}
		\caption{Depth Completion}
		\label{fig:fig_head_depth}
	\end{subfigure}
	\hspace{0.05\linewidth}
	\begin{subfigure}[t]{0.4\linewidth}
		\centering
		\includegraphics[width=\linewidth]{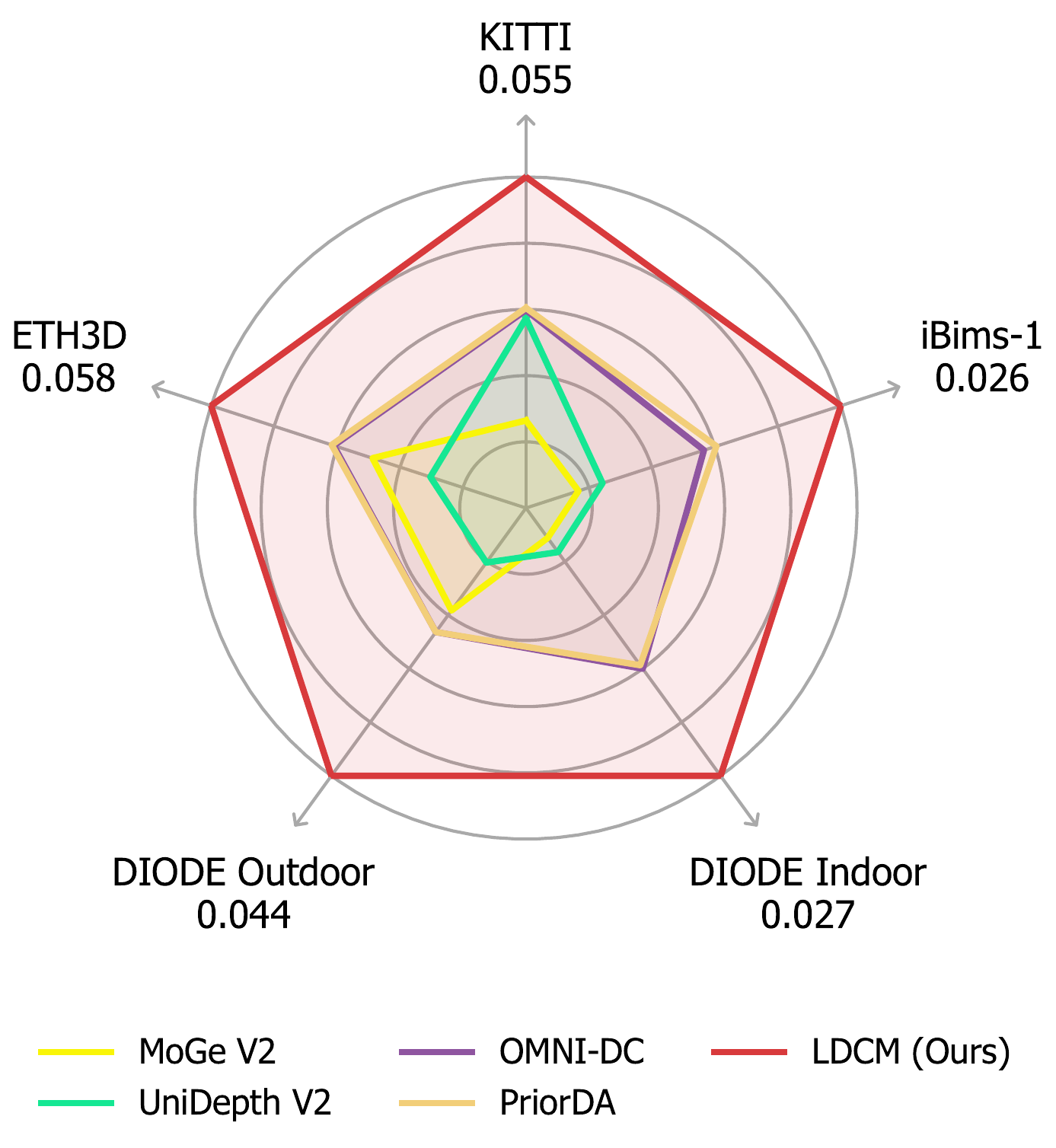}
		\caption{Point Map Estimation}
		\label{fig:fig_head_point}
	\end{subfigure}
	\caption{We present LDCM, a simple and effective model for depth completion.  Without complex architectural designs, LDCM achieves state-of-the-art performance in zero-shot depth completion and metric point map estimation. In the radar charts, larger areas indicate lower relative error (REL). LDCM ranks first across diverse datasets.}
	\label{fig:fig_head}
\end{figure}
\begin{abstract}
	This work presents the Large Depth Completion Model (LDCM), a simple, effective, and robust framework for single-view metric depth estimation with sparse observations. Without relying on complex architectural designs, LDCM generates metric-accurate dense depth maps using a transformer. It outperforms existing approaches across diverse datasets and sparse observations. We achieve this from two key perspectives: (1) leveraging existing monocular foundation models to improve the quality of sparse depth inputs, and (2) reformulating training objectives to better capture geometric structure and metric consistency. Specifically, a Poisson-based depth initialization strategy is first introduced to generate a uniform coarse dense depth map from diverse sparse observations, providing a strong structural prior for the network. Regarding the training objective, we replace the conventional depth head with a point map head that regresses per-pixel 3D coordinates in camera space, enabling the model to directly learn the underlying 3D scene structure instead of performing pixel-wise depth map restoration. Moreover, this design eliminates the need for camera intrinsic parameters, allowing LDCM to naturally produce metric-scaled 3D point maps. Extensive experiments demonstrate that LDCM consistently outperforms state-of-the-art methods across multiple benchmarks and varying sparsity levels in both depth completion and point map estimation, showcasing its effectiveness and strong generalization to unseen data distributions. 
	% Code and models are publicly available at \href{https://pkqbajng.github.io/ldcm/}{pkqbajng.github.io/ldcm/}.
\end{abstract}
\section{Introduction}
Dense depth maps are essential for applications in robotics~\cite{embodiedscan}, autonomous driving~\cite{bevbert}, and augmented reality~\cite{ar}. However, capturing dense and accurate depth data requires expensive active sensors such as LiDAR or structured light cameras, which are often limited by cost and hardware constraints. Thus, depth completion, which estimates a dense depth map from low-cost sparse depth observations and a corresponding RGB image, provides a cost-effective and efficient alternative.

While prior approaches~\cite{cspn, cspn2, rignet, rignetpp, nlspn, pointdc, bevdc} perform well on in-domain datasets such as NYUv2~\cite{nyuv2} and KITTI~\cite{kittidc}, they often fail to generalize to unseen environments and irregular sparse depth maps (\eg, Structure-from-Motion points with non-uniform density and large missing regions), limiting their real-world applicability. Driven by the success of foundation models trained on large-scale datasets~\cite{depthanything, depthanythingv2, metric3d, metric3dv2}, recent works~\cite{omnidc, g2md, priorda, spnet, pacgdc} have focused on architectural innovations and training with larger, more diverse data to improve robustness under domain shifts and varying sparsity. More recently, inspired by advances in natural language models~\cite{gpt4, qwen3}, prompt-based approaches~\cite{promptda, marigolddc, depthlab, depthprompting, priorda} treat the sparse depth map as a conditioning signal for transformer-based~\cite{depthanything, depthanythingv2} or diffusion-based~\cite{marigold, marigolddc, depthlab} depth foundation models, guiding the prediction toward metric-scale geometry. Despite their promising results, these methods fundamentally address depth completion as a depth restoration task, where the model learns to interpolate or denoise depth values conditioned on the sparse observation. This paradigm prioritizes local smoothness and texture-aware completion but lacks explicit 3D geometric reasoning, leading to unsatisfactory performance under severe domain shifts and highly irregular sparse depth maps.

In this work, we introduce the Large Depth Completion Model (LDCM), which produces dense, metric-accurate depth maps even from highly sparse and irregular observations. We achieve this by enhancing the input preprocessing pipeline and reformulating the training objective. To address the challenge of sparse and irregular depth maps, we leverage a monocular depth foundation model~\cite{depthanything, depthanythingv2} to enrich the geometric prior. Specifically, we construct a dense gradient field by combining a sparse depth map with relative depth cues predicted by the foundation model. We demonstrate that this hybrid gradient field serves as a proxy for solving a Poisson-based optimization problem, enabling the reconstruction of an initial coarse depth map that preserves fine geometric structures and exhibits metric-consistent depth values. Regarding the training objective, we replace the conventional depth regression head with a point map regression head, inspired by recent advances in 3D reconstruction~\cite{dust3r, mast3r, vggt, dens3r}. This reformulation explicitly encourages the network to predict metric-scale 3D coordinates, rather than focusing on pixel-wise restoration. The final depth map is obtained by extracting the z-component of the predicted point map, leading to more geometrically faithful and globally consistent predictions. Moreover, benefiting from this design, LDCM naturally predicts 3D point maps without requiring camera intrinsics, facilitating robust deployment in uncalibrated environments.

We perform extensive experiments to evaluate LDCM across six diverse benchmarks. The results demonstrate that our model surpasses all previous state-of-the-art methods in both depth completion and point map estimation, achieving top rankings across all tasks and metrics, as displayed in Fig.~\ref{fig:fig_head}. Our contribution can be summarized as follows:
\begin{itemize}
	\item We propose the Large Depth Completion Model (LDCM), which replaces the conventional depth regression head with a point map regression head to directly predict metric-scale 3D coordinates from a monocular image and sparse observations. This formulation facilitates more effective learning of metric-consistent 3D structures, leading to superior performance in dense depth completion.
	\item We introduce a Poisson-based coarse depth completion strategy that leverages relative depth cues from a monocular depth foundation model and sparse observations. This strategy generates high-quality initial depth maps, providing a geometrically faithful structural prior for subsequent feature learning.
	\item We demonstrate through extensive experiments that LDCM outperforms previous state-of-the-art methods in both depth completion and metric point map estimation across diverse benchmarks and varying sparsity levels, showcasing its robust generalization to unseen data.
\end{itemize}
\section{Related Work}
\textbf{Depth Completion.}
Depth completion aims to infer a dense depth map from a monocular image and a sparse depth map, which can be readily obtained from sources such as Structure-from-Motion~\cite{eth3d} or low-cost depth cameras~\cite{nyuv2}.  Recent deep learning-based approaches have achieved significant progress by proposing numerous spatial propagation network variants~\cite{spn, cspn, cspn2, nlspn, dyspn}  or exploiting visual structural guidance from images for guided restoration. To better exploit the 3D geometric information in sparse inputs, several 2D-3D joint depth completion approaches have also been proposed~\cite{pointdc, tpvd, tpvd2, bevdc}. Despite achieving impressive performance on single-domain datasets (\eg, NYUv2~\cite{nyuv2} and KITTI~\cite{kittidc}), these methods often struggle with cross-domain generalization, particularly when deployed in unseen environments and varying sparse observations. 

Inspired by the success of foundation models~\cite{sam, dinov2, depthanything, depthanythingv2, metric3d, metric3dv2, pacgdc} trained on large-scale datasets, recent works~\cite{omnidc, g2md, spnet, priorda} have focused on architectural innovations and training with larger, more diverse datasets to improve generalization. More recently, drawing inspiration from large language models~\cite{gpt4, qwen3}, prompt-based approaches~\cite{promptda, marigolddc, depthprompting, testpromptdc} have emerged that treat auxiliary priors as prompts to condition depth foundation models, effectively guiding predictions toward metric-scale outputs. PromptDA~\cite{promptda} introduces a compact prompt fusion architecture specifically designed for the DPT head~\cite{dpt}, enabling the integration of low-resolution depth cues. TestPromptDC~\cite{testpromptdc} presents a test-time prompt tuning method that adapts foundation models during inference without modifying their parameters, achieving sensor-specific depth scale adaptation while preserving foundational knowledge. MarigoldDC~\cite{marigolddc} prompts the sparse depth to a diffusion-based~\cite{marigold} foundation model. However, these methods fundamentally address depth completion as a depth restoration task, where the model learns to interpolate or denoise depth values conditioned on sparse inputs. The performance remains unsatisfactory under severe domain shifts and highly irregular sparse depth maps. In this work, we introduce a Poisson-based depth initialization module to effectively maximize the potential of depth foundation models to generate a coarse dense depth map, which serves as a strong structural prior for the following geometric feature learning. Besides, we reformulate the training objective as point maps, providing a more structurally faithful supervision for the network.

\textbf{Monocular Depth Estimation.}
A variety of monocular depth estimation foundation models~\cite{depthanything, depthanythingv2, unidepth, unidepthv2, metric3d, marigold, fe2e, jasmine} have been proposed. These models learn rich, generalizable priors from large-scale data and serve as strong backbones for downstream tasks such as stereo matching~\cite{foundationstereo, defomstereo, monster}, depth super-resolution~\cite{ducos}, depth completion~\cite{depthprompting, promptda, depthlab, marigolddc, priorda}, and autonomous driving~\cite{cgformer, voxdet, locc, voxformer, bevbert}. For instance, FoundationStereo~\cite{foundationstereo} introduces a side-tuning feature adapter that leverages monocular priors to bridge the sim-to-real domain gap. 
DuCos~\cite{ducos} treats foundation model outputs as structural priors for depth super-resolution (DSR) and seamlessly integrates them into a Lagrangian duality framework. 
PriorDA~\cite{priorda} employs a local weighted linear regression (LWLR) module~\cite{lwlr} to align the scale of relative depth with sparse observations, where the result is then refined by a structure-aware network to produce a dense depth map. However, this local alignment strategy often fails under highly sparse observations.
In contrast, we propose a novel Poisson-based initialization strategy to better exploit the potential of foundation models by enforcing gradient consistency constraints, yielding a significantly more geometrically coherent coarse depth map.

\textbf{Geometry Estimation Foundation Models.}
Point map~\cite{dust3r, vggt, moge, dens3r, more, pow3r}  representation has demonstrated strong potential for holistic scene understanding. Unlike depth maps, which encode 2.5D geometry tied to camera intrinsics, point maps explicitly model 3D structure. Several approaches~\cite{leres, unidepth, unidepthv2} decouple this task into depth prediction and camera parameter estimation. In contrast, DUSt3R~\cite{dust3r} bypasses explicit camera modeling by directly regressing a scale-invariant point map in an end-to-end fashion, with its successor Mast3R~\cite{mast3r} enabling metric-scale reconstruction. VGGT~\cite{vggt} introduces a feed-forward neural network capable of 3D reconstruction from one, a few, or even hundreds of input views of a scene. AnySplat~\cite{anysplat} extends VGGT~\cite{vggt} to support novel view synthesis from uncalibrated image collections. To facilitate single-view geometry learning, MoGe~\cite{moge, moge2} predicts an affine-invariant point map and recovers metric scale using a global scaling factor derived from contextual cues. More recently, several approaches~\cite{worldmirror, mapanything, pow3r} have introduced additional priors to enhance geometry estimation. Notably, Pow3R~\cite{pow3r} extends the DUSt3R~\cite{dust3r} paradigm by incorporating complementary modalities; however, it remains limited to relative geometry. In this work, we introduce point map representations for depth completion, enabling the model to directly learn the underlying 3D scene structure and produce metric quantities. Our concurrent work, MapAnything~\cite{mapanything}, also estimates metric 3D geometry from images and additional priors.
\section{Method}

\subsection{Overall Framework}
\begin{figure}[t]
	\centering
	\includegraphics[width=0.9\linewidth]{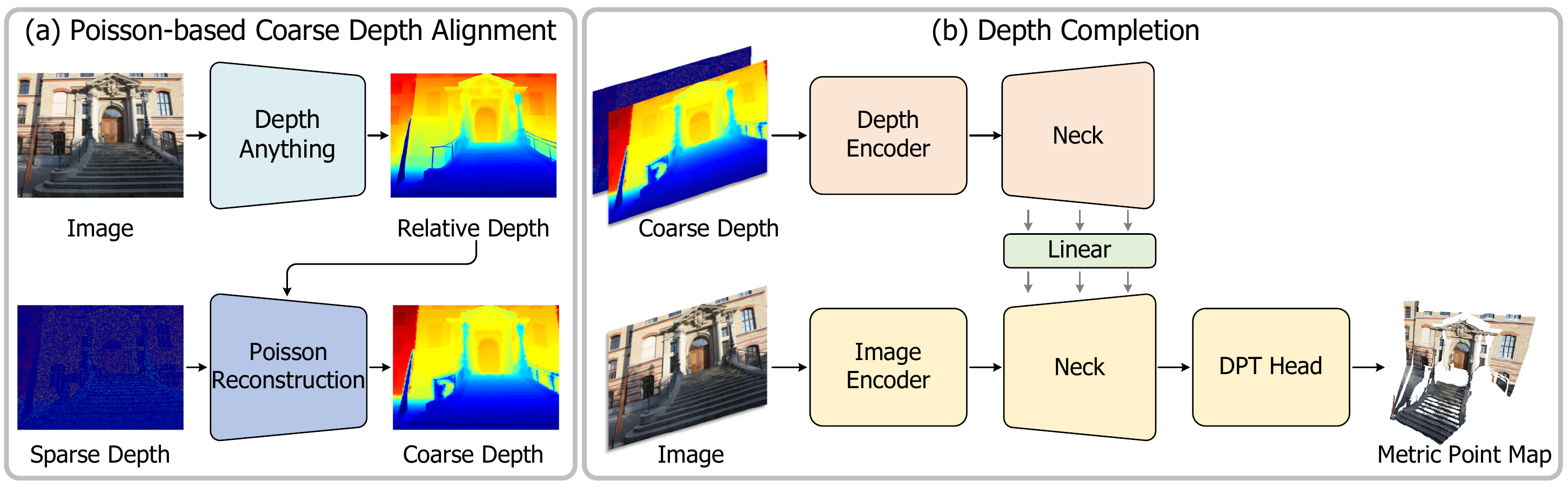}
	\caption{Schematics and detailed architecture of LDCM. Given a single image and sparse depth map, LDCM begins with a Poisson-based coarse depth alignment strategy. This process leverages a pretrained depth foundation model to generate an initial coarse depth map through gradient-domain optimization. This coarse depth, together with the input image, is then fed into the following point map prediction network to regress a dense, metric-scale 3D point map.}
	\label{fig:architecture}
\end{figure}

The framework of the proposed LDCM is illustrated in Fig.~\ref{fig:architecture}. Given an RGB image $\mathbf{I} \in \mathbb{R}^{H \times W \times 3}$ and a sparse depth map $\mathbf{S} \in \mathbb{R}^{H \times W}$, LDCM predicts a metric point map $\mathbf{P} \in \mathbb{R}^{H \times W \times 3}$ in camera space, from which the dense depth map is derived by extracting the z-channel component. The framework consists of two main stages. In the first stage, we harness the power of monocular depth foundation model to generate an initial coarse depth map $\mathbf{C}\in\mathbb{R}^{H\times W}$ via Poisson reconstruction. In the second stage, a ViT-based~\cite{vit} depth completion network takes the image $\mathbf{I}$ and the coarse depth $\mathbf{C}$ as input to predict the final metric 3D point map $\mathbf{P}$. The details of each stage are elaborated in the following sections.

\subsection{Coarse Depth Alignment}
\label{sec:coarse_align}
Different types of sparse depth priors exhibit distinct spatial distributions, ranging from random points and Structure-from-Motion keypoints to LiDAR-like structured sparsity, posing significant challenges for generalization. A straightforward approach involves direct interpolation of the sparse depth map~\cite{depthlab}; however, it often introduces severe artifacts due to the absence of strong geometric priors. With the advent of depth foundation models~\cite{midas, dpt}, which capture scene-level structure from large-scale training, leveraging them to provide robust geometric guidance has emerged as a promising direction.

To integrate sparse observations with foundation model predictions, we evaluate several coarse alignment strategies, including global affine alignment, local weighted linear regression (LWLR), and Poisson-based optimization. While the former two offer simple parametric alignment, they exhibit critical limitations.  Global affine alignment assumes a uniform scale and shift across the entire image, making it unable to recover per-pixel metric values. LWLR improves spatial adaptivity by fitting local models, but its performance is highly sensitive to the distribution and density of sparse depth maps. In contrast, Poisson-based optimization formulates the alignment as a gradient-field reconstruction problem, demonstrating superior geometric coherence and metric accuracy across diverse sparse observations. Therefore, we adopt Poisson reconstruction in the first stage of LDCM to generate the coarse depth map $\mathbf{C}$.

Specifically, given a sparse depth input $\mathbf{S}$ and relative depth cues $\mathbf{D}_r$ from a foundation model, we aim to generate a coarse dense depth map $\mathbf{C}$ that aligns with the geometric structure of $\mathbf{D}_r$, while preserving the observed values in $\mathbf{S}$. The problem can be formulated as minimizing the following optimization objective:
\begin{equation}
	\mathbf{C} = \arg\min_{\mathbf{D}} \left( \sum_{i} \|\nabla \log \mathbf{D}_{i} - \mathbf{G}_{i}\|^2 + \lambda \sum_{i \in \Omega} (\mathbf{D}_{i} - \mathbf{S}_{i})^2 \right),
\end{equation}
where $\mathbf{G}$ is a target log-gradient field that encodes structural fidelity and metric consistency, $\Omega$ denotes the set of valid sparse depth points, and $\lambda$ balances two terms.
A naive choice is $\mathbf{G} = \nabla \log \mathbf{D}_r$, but this ignores the unknown scale and shift of relative depth and may lead to misaligned gradients in metric space. Instead, we construct a more informed target by incorporating metric priors from sparse observations.
Let $(\alpha, \beta)$ be the global affine transformation that best aligns $\mathbf{D}_r$ with $\mathbf{S}$:
\begin{equation}
	(\alpha, \beta) = \arg\min_{\alpha', \beta'} \sum_{i \in \Omega} \left( \mathbf{S}_{i} - \alpha' \cdot (\mathbf{D}_r)_{i} - \beta' \right)^2,\label{eq:global_align}
\end{equation}
and define $\gamma = \beta / \alpha$. We then set:
\begin{equation}
	\mathbf{G} = \nabla \log(\mathbf{D}_r + \gamma).
\end{equation}

This choice is motivated by the fact that during training, the relative depth ground truth $\mathbf{D}_r$ is derived from the metric ground truth $\mathbf{D}^*$ via an affine transformation: $\mathbf{D}_r = (\mathbf{D}^* - \beta)/\alpha$. While this ideal relationship may not strictly hold for the predicted $\mathbf{D}_r$ during inference, introducing a shift $\gamma$ helps align its gradient structure with the metric space. Empirically, $\nabla \log(\mathbf{D}_r + \gamma)$ serves as a robust proxy for the target log-gradient field, preserving fine geometric details while being anchored to the metric scale through sparse inputs. Thus, the final formulation becomes:
\begin{equation}
	\mathbf{C} = \arg\min_{\mathbf{D}} \left( \sum_{i} \|\nabla \log \mathbf{D}_{i} - \nabla \log(\mathbf{D}_r + \gamma)_{i}\|^2 + \lambda \sum_{i \in \Omega} (\mathbf{D}_{i} - \mathbf{S}_{i})^2 \right),
\end{equation}
which can be solved through the conjugate gradient method~\cite{conjugate}. In this formulation, each sparse point anchors the global energy, and due to the nature of gradient-domain reconstruction, its influence propagates across the entire image via the structural constraints encoded in the gradient field.

\subsection{Depth Completion Network}
The architecture is illustrated in Fig.~\ref{fig:architecture}(b). We employ dual encoders to extract features from the coarse depth map $\mathbf{C}$ and the RGB image, respectively. Features are fused using the prompt fusion block~\cite{promptda}.  For the final output, instead of regressing a depth map, we replace the standard depth regression head with a point map head that directly predicts per-pixel 3D coordinates $\mathbf{P}$. This enables the model to learn the underlying 3D scene structure holistically, rather than performing pixel-wise depth restoration. Ablation studies demonstrate that this design leads to better accuracy. Moreover, thanks to this end-to-end formulation, the model naturally produces metric 3D point maps, facilitating robust deployment in uncalibrated environments.

\subsection{Training}
\textbf{Training Losses.} We train the LDCM using three complementary losses on the predicted 3D point map $\mathbf{P}$, with the ground-truth metric point map denoted as $\hat{\mathbf{P}}$.
\begin{equation}
	\mathcal{L} = \mathcal{L}_{\text{global}} + \lambda_{\text{local}} \mathcal{L}_{\text{local}} + \lambda_{\text{normal}} \mathcal{L}_{\text{normal}},
\end{equation}
where the individual terms are defined as follows. The global point map loss enforces overall structural consistency:
\begin{equation}
	\mathcal{L}_{\text{global}} = \sum_{i \in \mathcal{M}} \frac{1}{\hat{\mathbf{D}}_i} \|\mathbf{P}_i - \hat{\mathbf{P}}_i\|_1,
\end{equation}
where $\mathcal{M}$ denotes the region of valid ground-truth. The local point map loss captures fine-grained geometry by operating in 3D neighborhoods. Following~\cite{moge}, we sample anchor points and define spherical regions $\mathcal{S}_j$ in 3D space:
\begin{equation}
	\mathcal{L}_{\text{local}} = \sum_{j \in \mathcal{H}_a} \sum_{i \in \mathcal{S}_j} \frac{1}{\hat{\mathbf{D}}_i} \|\mathbf{P}_i - \hat{\mathbf{P}}_i\|_1.
\end{equation}
This encourages local coherence independent of image perspective. The normal loss promotes surface smoothness and alignment:
\begin{equation}
	\mathcal{L}_{\text{normal}} = \sum_{i \in \mathcal{M}} \arccos\left( \frac{\mathbf{N}_i^\top \hat{\mathbf{N}}_i}{\|\mathbf{N}_i\| \|\hat{\mathbf{N}}_i\|} \right),
\end{equation}
where $\mathbf{N}_i$ and $\hat{\mathbf{N}}_i$ are surface normals estimated from $\mathbf{P}$ and $\hat{\mathbf{P}}$, respectively.

\textbf{Implementation Details.}
We train the LDCM on 11 public RGB-D datasets~\cite{hypersim, tartanair, irs, pointodyssey, urbansyn, synscapes, matrixcity, lightwheelocc, mvssynth, synthia, scannetpp}, approximately 2.7 million samples. The combined data covers diverse indoor and outdoor scenes; further details are provided in the \textit{suppl. material}. 

LDCM uses a ViT-B~\cite{vit} pretrained with DINOv2~\cite{dinov2} as the image encoder. For coarse depth alignment, we use DepthAnythingV2-S~\cite{depthanythingv2} as the foundation model. Training runs for $200\,\text{K}$ iterations using the AdamW optimizer~\cite{adamw} with a cosine learning rate schedule and linear warmup over the first 5\% of iterations. The peak learning rates are $1 \times 10^{-5}$ for the encoder and $1 \times 10^{-4}$ for all other layers. We use a global batch size of $128$, with mini-batches sampling an approximately equal number of images from each dataset. During training, images are resized such that their aspect ratios range uniformly from $1:2$ to $2:1$, and total pixel counts fall between $250\,\text{K}$ and $500\,\text{K}$. Data augmentation includes random cropping, color jittering, Gaussian blur, JPEG compression-decompression, and perspective-aware cropping to align the principal point with the image center. Sparse depth inputs are synthetically generated by subsampling dense ground-truth depth maps with varying patterns, following the protocol of OMNI-DC~\cite{omnidc}. The training is conducted on 16 H20 GPUs and takes approximately six days to complete.

\section{Experiments}

\subsection{Quantitative Evaluations}
We evaluate the zero-shot performance of LDCM and compare it with several state-of-the-art approaches for depth completion~\cite{g2md, omnidc, spnet, promptda, priorda}, monocular depth estimation~\cite{depthanything, depthanythingv2, vggt, depthpro}, and monocular point map estimation~\cite{unidepth, unidepthv2, moge, moge2, worldmirror, pow3r, mapanything}. Additional details on the compared approaches and evaluation protocols are provided in the \textit{suppl. material}. As demonstrated in the experiments, LDCM achieves superior performance across multiple benchmarks.

\begin{table}[t]
	\caption{\textbf{Quantitative comparison of depth completion methods on benchmark datasets.} All methods are evaluated under zero-shot settings. Methods marked with $\dagger$ produce relative depth, and metric depth is recovered by optimizing global scale and shift via least squares regression using the \textit{same} sparse depth prior. Methods marked with $\ddagger$ use dataset-specific configurations for indoor and outdoor scenes, respectively. The \colorbox{best}{best} and \colorbox{best2}{second-best} results are highlighted.}
	\resizebox{\columnwidth}{!}{%
		\begin{tabular}{cccccccccccccccc}
			\specialrule{1.5pt}{0.8ex}{0.8ex}
			\multirow{2}{*}{Method} & \multicolumn{5}{c}{KITTI}                & \multicolumn{5}{c}{iBims-1}                 & \multicolumn{5}{c}{DIODE Indoor}              \\\cmidrule(l){2-6} \cmidrule(l){7-11} \cmidrule(l){12-16} 
			& RMSE$\downarrow$ & MAE$\downarrow$   & REL$\downarrow$   & $\delta_1\uparrow$    & Rk.$\downarrow$ & RMSE$\downarrow$ & MAE$\downarrow$   & REL$\downarrow$   & $\delta_1\uparrow$  & Rk.$\downarrow$ & RMSE$\downarrow$ & MAE$\downarrow$   & REL$\downarrow$   & $\delta_1\uparrow$  & Rk.$\downarrow$ \\\midrule
			DepthPro & 4.149 & 2.763 & 0.178 & 0.731 & 13.432 & 0.605 & 0.503 & 0.156 & 0.829 & 13.750 & 0.837 & 0.702 & 0.193 & 0.668 & 14.114 \\
			UniDepth V1 & 3.335 & 2.010 & 0.118 & 0.938 & 8.636 & 1.166 & 1.082 & 0.370 & 0.236 & 16.000 & 0.939 & 0.840 & 0.158 & 0.779 & 13.523 \\
			UniDepth V2 & 3.150 & 1.598 & 0.090 & 0.960 & 6.500 & 0.446 & 0.321 & 0.100 & 0.935 & 11.932 & 0.811 & 0.678 & 0.165 & 0.681 & 13.023 \\
			DepthAnythingV2$\dagger$ & 4.007 & 1.890 & 0.092 & 0.916 & 9.091 & 0.349 & 0.179 & 0.043 & 0.975 & 8.295 & 0.386 & 0.189 & 0.045 & 0.976 & 7.295 \\
			VGGT$\dagger$ & 4.219 & 2.518 & 0.158 & 0.783 & 12.909 & 0.348 & 0.194 & 0.053 & 0.957 & 10.318 & 0.425 & 0.294 & 0.096 & 0.920 & 10.773 \\
			MoGe V1$\dagger$ & 3.050 & 1.821 & 0.125 & 0.887 & 8.568 & 0.238 & 0.120 & 0.035 & 0.981 & 6.045 & 0.272 & 0.175 & 0.064 & 0.950 & 7.386 \\
			MoGe V2 & 4.617 & 3.366 & 0.213 & 0.458 & 15.182 & 0.633 & 0.540 & 0.156 & 0.707 & 14.500 & 1.064 & 0.938 & 0.235 & 0.433 & 15.841 \\
			G2-MonoDepth$\ddagger$ & 2.638 & 0.964 & 0.054 & 0.949 & 5.295 & 0.227 & 0.094 & 0.028 & 0.973 & 5.409 & 0.298 & 0.198 & 0.067 & 0.879 & 6.341 \\
			OMNI-DC & \colorbox{best2}{2.302} & 0.760 & 0.042 & 0.963 & 3.045 & 0.192 & 0.063 & 0.018 & 0.982 & 2.932 & 0.141 & 0.064 & 0.022 & 0.968 & \colorbox{best2}{2.932} \\
			PriorDA & 2.364 & 0.861 & 0.044 & \colorbox{best2}{0.971} & 4.159 & \colorbox{best2}{0.176} & 0.065 & 0.018 & \colorbox{best2}{0.990} & 3.477 & \colorbox{best2}{0.093} & \colorbox{best2}{0.037} & \colorbox{best2}{0.012} & \colorbox{best}{0.994} & 3.023 \\
			SPNet$\ddagger$ & 2.365 & \colorbox{best2}{0.757} & \colorbox{best2}{0.041} & 0.966 & \colorbox{best2}{3.000} & 0.189 & \colorbox{best2}{0.059} & \colorbox{best2}{0.016} & 0.987 & \colorbox{best2}{2.659} & 0.157 & 0.078 & 0.028 & 0.954 & 3.273 \\
			PromptDA & 3.040 & 1.261 & 0.067 & 0.946 & 6.545 & 0.249 & 0.116 & 0.033 & 0.975 & 6.091 & 0.203 & 0.115 & 0.037 & 0.965 & 6.068 \\
			WorldMirror$\dagger$ & 4.439 & 2.432 & 0.142 & 0.824 & 11.818 & 0.352 & 0.192 & 0.051 & 0.963 & 9.205 & 0.386 & 0.243 & 0.084 & 0.941 & 9.364 \\
			MapAnything & 12.974 & 6.784 & 0.350 & 0.588 & 15.750 & 0.968 & 0.374 & 0.104 & 0.909 & 13.295 & 0.909 & 0.458 & 0.104 & 0.899 & 11.000 \\
			Pow3R$\dagger$ & 3.515 & 2.096 & 0.141 & 0.832 & 10.750 & 0.338 & 0.183 & 0.049 & 0.965 & 9.091 & 0.353 & 0.240 & 0.078 & 0.943 & 9.000 \\\midrule
			LDCM (Ours) & \colorbox{best}{1.911} & \colorbox{best}{0.537} & \colorbox{best}{0.026} & \colorbox{best}{0.983} & \colorbox{best}{1.068} & \colorbox{best}{0.161} & \colorbox{best}{0.044} & \colorbox{best}{0.012} & \colorbox{best}{0.991} & \colorbox{best}{1.659} & \colorbox{best}{0.084} & \colorbox{best}{0.025} & \colorbox{best}{0.008} & \colorbox{best2}{0.993} & \colorbox{best}{1.545} \\\specialrule{1pt}{0.4ex}{0.4ex}
			\multirow{2}{*}{Method} & \multicolumn{5}{c}{DIODE Outdoor}        & \multicolumn{5}{c}{ETH3D}                  & \multicolumn{5}{c}{Average}              \\\cmidrule(l){2-6} \cmidrule(l){7-11} \cmidrule(l){12-16} 
			& RMSE$\downarrow$ & MAE$\downarrow$ & REL$\downarrow$   & $\delta_1\uparrow$ & Rk.$\downarrow$ & RMSE$\downarrow$ & MAE$\downarrow$   & REL$\downarrow$ & $\delta_1\uparrow$ & Rk.$\downarrow$ & RMSE$\downarrow$ & MAE$\downarrow$ & REL$\downarrow$ & $\delta_1\uparrow$ & Rk.$\downarrow$ \\\midrule
			DepthPro & 9.539 & 7.635 & 0.403 & 0.177 & 14.636 & 3.199 & 2.562 & 0.302 & 0.477 & 15.023 & 3.666 & 2.833 & 0.246 & 0.576 & 14.191 \\
			UniDepth V1 & 5.782 & 3.841 & 0.189 & 0.661 & 11.795 & 3.482 & 3.170 & 0.579 & 0.116 & 15.728 & 2.941 & 2.189 & 0.283 & 0.546 & 13.136 \\
			UniDepth V2 & 11.145 & 8.936 & 0.515 & 0.526 & 15.250 & 1.630 & 1.169 & 0.200 & 0.726 & 13.387 & 3.436 & 2.540 & 0.214 & 0.766 & 12.018 \\
			DepthAnythingV2$\dagger$ & 5.940 & 2.777 & 0.124 & 0.869 & 8.659 & 2.091 & 0.424 & 0.049 & 0.979 & 9.477 & 2.555 & 1.092 & 0.071 & 0.943 & 8.563 \\
			VGGT$\dagger$ & 4.898 & 2.893 & 0.237 & 0.772 & 10.591 & 0.540 & 0.317 & 0.060 & 0.949 & 9.103 & 2.086 & 1.243 & 0.121 & 0.876 & 10.739 \\
			MoGe V1$\dagger$ & 10.576 & 8.340 & 0.406 & 0.599 & 14.250 & 1.651 & 0.550 & 0.082 & 0.943 & 8.750 & 3.157 & 2.201 & 0.142 & 0.872 & 9.000 \\
			MoGe V2 & 4.807 & 3.352 & 0.182 & 0.680 & 10.477 & 0.847 & 0.619 & 0.114 & 0.839 & 11.784 & 2.394 & 1.763 & 0.180 & 0.623 & 13.557 \\
			G2-MonoDepth$\ddagger$ & 2.393 & 0.875 & 0.062 & 0.938 & 4.682 & 0.428 & 0.177 & 0.034 & 0.969 & 5.603 & 1.197 & 0.462 & 0.049 & 0.942 & 5.466 \\
			OMNI-DC & 2.322 & 0.726 & 0.049 & 0.955 & 3.341 & 0.290 & \colorbox{best2}{0.100} & \colorbox{best2}{0.016} & 0.987 & \colorbox{best2}{2.932} & 1.049 & 0.343 & 0.029 & 0.971 & 3.036 \\
			PriorDA & 2.310 & 0.858 & 0.051 & 0.957 & 3.932 & \colorbox{best2}{0.274} & 0.105 & 0.017 & \colorbox{best2}{0.990} & 3.443 & \colorbox{best2}{1.043} & 0.385 & \colorbox{best2}{0.028} & \colorbox{best2}{0.980} & 3.607 \\
			SPNet$\ddagger$ & \colorbox{best2}{2.111} & \colorbox{best2}{0.658} & \colorbox{best2}{0.048} & \colorbox{best2}{0.959} & \colorbox{best2}{2.114} & 0.419 & 0.119 & 0.019 & 0.986 & 3.625 & 1.048 & \colorbox{best2}{0.334} & 0.030 & 0.970 & \colorbox{best2}{2.934} \\
			PromptDA & 3.604 & 1.561 & 0.087 & 0.912 & 6.182 & 0.644 & 0.276 & 0.041 & 0.967 & 7.102 & 1.548 & 0.666 & 0.053 & 0.953 & 6.398 \\
			WorldMirror$\dagger$ & 4.464 & 2.317 & 0.151 & 0.828 & 8.045 & 0.524 & 0.302 & 0.051 & 0.962 & 7.761 & 2.033 & 1.097 & 0.096 & 0.904 & 9.239 \\
			MapAnything & 7.675 & 3.891 & 0.219 & 0.731 & 11.318 & 1.952 & 0.711 & 0.108 & 0.904 & 12.523 & 4.896 & 2.444 & 0.177 & 0.806 & 12.777 \\
			Pow3R$\dagger$ & 3.682 & 2.068 & 0.169 & 0.840 & 7.568 & 0.480 & 0.273 & 0.048 & 0.964 & 7.545 & 1.674 & 0.972 & 0.097 & 0.909 & 8.791 \\\midrule
			LDCM (Ours) & \colorbox{best}{1.969} & \colorbox{best}{0.529} & \colorbox{best}{0.031} & \colorbox{best}{0.970} & \colorbox{best}{1.568} & \colorbox{best}{0.187} & \colorbox{best}{0.048} & \colorbox{best}{0.008} & \colorbox{best}{0.997} & \colorbox{best}{1.148} & \colorbox{best}{0.862} & \colorbox{best}{0.237} & \colorbox{best}{0.017} & \colorbox{best}{0.987} & \colorbox{best}{1.398} \\
			\specialrule{1.5pt}{0.8ex}{0.8ex}
			\end{tabular}%
		}
	\label{tab:synthetic_depth}
	\vspace{-4mm}
\end{table}

\textbf{Depth completion.} We evaluate depth completion on KITTI~\cite{kittidc}, ETH3D~\cite{eth3d}, iBims-1~\cite{ibims}, and DIODE~\cite{diode}, covering both indoor and outdoor scenarios. To assess robustness under diverse sparse sampling patterns, we synthesize sparse depth inputs using the following strategies:
\begin{itemize}
	\item \textbf{Noisy random sampling}: uniformly sampled points at varying densities (e.g., 1\%, 3\%, 5\%, 10\%), with mild noise simulation;
	\item \textbf{Keypoint-based sampling}: depth values extracted at SIFT or ORB keypoints;
	\item \textbf{LiDAR-simulated sampling}: synthetic LiDAR scans with varying numbers of vertical beams (\eg, 64, 32, 16 lines).
\end{itemize}
On KITTI, the simulation is applied to raw single-frame LiDAR measurements~\cite{omnidc, lrru}. For all other datasets, they are generated from dense ground-truth depth maps.  We evaluate the predicted depth maps using standard metrics: Root Mean Squared Error (RMSE), Mean Absolute Error (MAE), Relative Error (REL), and the accuracy threshold $\delta_1$.  For methods that produce relative depth maps~\cite{vggt, depthanythingv2, moge}, we recover the global scale and shift via least squares regression using the sparse depth prior. Table~\ref{tab:synthetic_depth} reports the average RMSE, MAE, REL, and $\delta_1$ across all synthetic patterns per dataset, along with the mean ranking over competing methods. As shown in the table, LDCM achieves state-of-the-art performance. Notably, it maintains high accuracy even under extreme sparsity, demonstrating strong robustness and generalization across diverse sparse input configurations. 

\begin{table}[h]
	\caption{\textbf{Quantitative comparison of point map estimation methods on benchmark datasets.} All methods are evaluated under zero-shot settings. The \colorbox{best}{best} and \colorbox{best2}{second-best} results are highlighted.}
	\resizebox{\columnwidth}{!}{%
		\begin{tabular}{cccccccccccccccc}
			\specialrule{1.5pt}{0.8ex}{0.8ex}
			\multirow{2}{*}{Method} & \multicolumn{5}{c}{KITTI}                & \multicolumn{5}{c}{iBims-1}                 & \multicolumn{5}{c}{DIODE Indoor}              \\\cmidrule(l){2-6} \cmidrule(l){7-11} \cmidrule(l){12-16} 
			& $\text{MAE}^{p}\downarrow$ & $\text{RMSE}^{p}\downarrow$   & $\text{REL}^{p}\downarrow$   & $\delta_1^{p}\uparrow$    & Rk.$\downarrow$ & $\text{MAE}^{p}\downarrow$  & $\text{RMSE}^{p}\downarrow$  & $\text{REL}^{p}\downarrow$    & $\delta_1^{p}\uparrow$    & Rk.$\downarrow$ & $\text{MAE}^{p}\downarrow$ & $\text{RMSE}^{p}\downarrow$  & $\text{REL}^{p}\downarrow$   & $\delta_1^{p}\uparrow$    & Rk.$\downarrow$\\\midrule
			UniDepth V1 & 2.207 & 3.540 & 0.120 & 0.954 & 6.773 & 1.154 & 1.239 & 0.370 & 0.239 & 9.000 & 0.911 & 1.017 & 0.159 & 0.779 & 7.318 \\
			UniDepth V2 & 1.813 & 3.540 & 0.096 & 0.961 & 5.409 & 0.365 & 0.489 & 0.107 & 0.932 & 6.909 & 0.730 & 0.872 & 0.164 & 0.694 & 7.273 \\
			MoGe V2 & 3.536 & 4.899 & 0.208 & 0.484 & 9.000 & 0.574 & 0.667 & 0.156 & 0.740 & 8.000 & 1.048 & 1.185 & 0.242 & 0.410 & 8.955 \\
			G2-MonoDepth & 1.669 & 3.118 & 0.098 & 0.946 & 4.841 & 0.186 & 0.287 & 0.052 & 0.972 & 4.750 & 0.305 & 0.401 & 0.087 & 0.875 & 4.841 \\
			OMNI-DC & 1.542 & \colorbox{best2}{2.828} & 0.092 & 0.960 & 3.409 & 0.164 & 0.256 & 0.046 & 0.980 & 3.341 & 0.174 & 0.241 & 0.045 & 0.967 & 2.977 \\
			PriorDA & 1.573 & 2.836 & 0.091 & \colorbox{best2}{0.965} & 4.341 & 0.159 & 0.240 & 0.043 & \colorbox{best2}{0.989} & 3.500 & \colorbox{best2}{0.140} & \colorbox{best2}{0.190} & \colorbox{best2}{0.034} & \colorbox{best}{0.994} & 2.909 \\
			SPNet & \colorbox{best2}{1.507} & 2.881 & \colorbox{best2}{0.089} & 0.964 & \colorbox{best2}{3.068} & \colorbox{best2}{0.152} & \colorbox{best2}{0.239} & \colorbox{best2}{0.042} & 0.988 & \colorbox{best2}{2.455} & 0.172 & 0.236 & 0.046 & 0.963 & \colorbox{best2}{2.636} \\
			PromptDA & 1.933 & 3.612 & 0.110 & 0.938 & 6.659 & 0.199 & 0.309 & 0.054 & 0.975 & 5.545 & 0.204 & 0.301 & 0.056 & 0.963 & 5.523 \\\midrule
			LDCM (Ours) & \colorbox{best}{\textbf{1.027}} & \colorbox{best}{\textbf{2.308}} & \colorbox{best}{\textbf{0.055}} & \colorbox{best}{\textbf{0.982}} & \colorbox{best}{\textbf{1.045}} & \colorbox{best}{\textbf{0.092}} & \colorbox{best}{\textbf{0.194}} & \colorbox{best}{\textbf{0.026}} & \colorbox{best}{\textbf{0.992}} & \colorbox{best}{\textbf{1.000}} & \colorbox{best}{\textbf{0.127}} & \colorbox{best}{\textbf{0.179}} & \colorbox{best}{\textbf{0.027}} & \colorbox{best2}{\textbf{0.992}} & \colorbox{best}{\textbf{1.159}} \\\specialrule{1pt}{0.4ex}{0.4ex}
			\multirow{2}{*}{Method} & \multicolumn{5}{c}{DIODE Outdoor}        & \multicolumn{5}{c}{ETH3D}                  & \multicolumn{5}{c}{Average}              \\\cmidrule(l){2-6} \cmidrule(l){7-11} \cmidrule(l){12-16} 
			& $\text{MAE}^{p}\downarrow$ & $\text{RMSE}^{p}\downarrow$   & $\text{REL}^{p}\downarrow$   & $\delta_1^{p}\uparrow$    & Rk.$\downarrow$ & $\text{MAE}^{p}\downarrow$  & $\text{RMSE}^{p}\downarrow$  & $\text{REL}^{p}\downarrow$    & $\delta_1^{p}\uparrow$    & Rk.$\downarrow$ & $\text{MAE}^{p}\downarrow$ & $\text{RMSE}^{p}\downarrow$  & $\text{REL}^{p}\downarrow$   & $\delta_1^{p}\uparrow$    & Rk.$\downarrow$\\\midrule
			UniDepth V1 & 4.653 & 5.100 & 0.461 & 0.145 & 9.000 & 3.541 & 3.875 & 0.551 & 0.106 & 9.000 & 2.493 & 2.954 & 0.332 & 0.445 & 8.218 \\
			UniDepth V2 & 1.879 & 2.844 & 0.216 & 0.712 & 8.000 & 1.252 & 1.785 & 0.191 & 0.769 & 8.000 & 1.208 & 1.906 & 0.155 & 0.814 & 7.118 \\
			MoGe V2 & 0.931 & 1.206 & 0.115 & 0.890 & 5.977 & 0.716 & 0.913 & 0.119 & 0.865 & 6.409 & 1.361 & 1.774 & 0.168 & 0.678 & 7.668 \\
			G2-MonoDepth & 0.794 & 1.129 & 0.109 & 0.891 & 4.864 & 0.603 & 0.826 & 0.105 & 0.911 & 5.160 & 0.711 & 1.152 & 0.090 & 0.919 & 4.891 \\
			OMNI-DC & 0.714 & 0.946 & \colorbox{best2}{0.095} & 0.915 & \colorbox{best2}{2.795} & 0.550 & 0.710 & 0.095 & 0.929 & \colorbox{best2}{3.284} & 0.629 & 0.996 & 0.075 & 0.950 & \colorbox{best2}{3.161} \\
			PriorDA & \colorbox{best2}{0.698} & \colorbox{best2}{0.908} & \colorbox{best2}{0.095} & \colorbox{best2}{0.919} & 3.295 & \colorbox{best2}{0.538} & \colorbox{best2}{0.682} & 0.094 & \colorbox{best2}{0.936} & 3.352 & \colorbox{best2}{0.622} & \colorbox{best2}{0.971} & \colorbox{best2}{0.071} & \colorbox{best2}{0.961} & 3.479 \\
			SPNet & 0.733 & 1.243 & 0.100 & 0.914 & 3.932 & 0.557 & 0.859 & 0.096 & 0.931 & 3.796 & 0.624 & 1.092 & 0.075 & 0.952 & 3.177 \\
			PromptDA & 0.824 & 1.422 & 0.100 & 0.911 & 5.591 & 0.592 & 0.950 & \colorbox{best2}{0.093} & 0.932 & 4.159 & 0.750 & 1.319 & 0.083 & 0.944 & 5.495 \\\midrule
			LDCM (Ours) & \colorbox{best}{\textbf{0.427}} & \colorbox{best}{\textbf{0.580}} & \colorbox{best}{\textbf{0.044}} & \colorbox{best}{\textbf{0.995}} & \colorbox{best}{\textbf{1.000}} & \colorbox{best}{\textbf{0.347}} & \colorbox{best}{\textbf{0.456}} & \colorbox{best}{\textbf{0.058}} & \colorbox{best}{\textbf{0.996}} & \colorbox{best}{\textbf{1.000}} & \colorbox{best}{\textbf{0.404}} & \colorbox{best}{\textbf{0.743}} & \colorbox{best}{\textbf{0.042}} & \colorbox{best}{\textbf{0.991}} & \colorbox{best}{\textbf{1.041}} \\
			\specialrule{1.5pt}{0.8ex}{0.8ex}
		\end{tabular}%
	}
	\label{tab:synthetic_point}
	\vspace{-4mm}
\end{table}

\textbf{Point map estimation.}
We adopt the same benchmarks used for depth completion to evaluate monocular point map estimation. The predicted point maps are evaluated using point-wise metrics: $\text{RMSE}^{p}$, $\text{MAE}^{p}$, $\text{REL}^{p}$ and $\delta_1^{p}$.  Table~\ref{tab:synthetic_point} reports the average performance across all synthetic patterns per dataset for each metric. For depth completion methods, we use the camera intrinsics from UniDepth V2~\cite{unidepthv2} to back-project the completed depth maps into 3D point maps. As shown in the table, LDCM consistently outperforms all competing methods, achieving the best results across all datasets and metrics.

\textbf{Affine-invariant point map estimation.} We adopt the same benchmarks to evaluate monocular affine-invariant point map estimation. Following MoGe~\cite{moge}, we resolve the scale and shift of the predicted point map using the proposed ROE solver to align it with the ground truth. Table~\ref{tab:synthetic_affine_invariant_point} reports the average performance in terms of $\text{REL}^{p}$ and $\delta_1^{p}$. As shown in the table, our method achieves superior performance compared to baseline approaches and outperforms state-of-the-art relative geometry estimation methods, including VGGT~\cite{vggt} and WorldMirror~\cite{worldmirror}. This demonstrates that our model preserves—rather than compromises—the accuracy of relative geometry estimation.

\begin{table}[t]
	\caption{\textbf{Quantitative comparison of affine-invariant point map estimation methods on benchmark datasets.} All methods are evaluated under zero-shot settings. The \colorbox{best}{best} and \colorbox{best2}{second-best} results are highlighted.}
	\centerfirst
	\resizebox{\columnwidth}{!}{%
		\begin{tabular}{cccccccccc}
			\specialrule{1.5pt}{0.8ex}{0.8ex}
			\multirow{2}{*}{Method} & \multicolumn{3}{c}{KITTI}                & \multicolumn{3}{c}{iBims-1}                 & \multicolumn{3}{c}{DIODE Indoor}              \\\cmidrule(l){2-4} \cmidrule(l){5-7} \cmidrule(l){8-10} 
			& $\text{REL}^{p}\downarrow$   & $\delta_1^{p}\uparrow$    & Rk.$\downarrow$ & $\text{REL}^{p}\downarrow$    & $\delta_1^{p}\uparrow$    & Rk.$\downarrow$  & $\text{REL}^{p}\downarrow$   & $\delta_1^{p}\uparrow$    & Rk.$\downarrow$\\\midrule
			VGGT & 0.147 & 0.823 & 4.500 & 0.048 & 0.967 & 3.909 & 0.107 & 0.926 & 4.636 \\
			MoGe V2 & \colorbox{best2}{0.056} & \colorbox{best2}{0.968} & \colorbox{best2}{1.909} & 0.046 & \colorbox{best2}{0.972} & \colorbox{best2}{2.455} & \colorbox{best2}{0.052} & \colorbox{best2}{0.972} & \colorbox{best2}{1.955} \\
			WorldMirror & 0.108 & 0.886 & 3.136 & \colorbox{best2}{0.044} & 0.965 & 2.864 & 0.073 & 0.953 & 3.091 \\
			MapAnything & 0.366 & 0.344 & 6.000 & 0.233 & 0.611 & 6.000 & 0.172 & 0.758 & 6.000 \\
			Pow3R & 0.152 & 0.850 & 4.318 & 0.064 & 0.965 & 4.318 & 0.108 & 0.947 & 4.273 \\\midrule
			LDCM (Ours) & \colorbox{best}{0.039} & \colorbox{best}{0.983} & \colorbox{best}{1.091} & \colorbox{best}{0.017} & \colorbox{best}{0.992} & \colorbox{best}{1.000} & \colorbox{best}{0.014} & \colorbox{best}{0.995} & \colorbox{best}{1.000} \\\specialrule{1pt}{0.4ex}{0.4ex}
			\multirow{2}{*}{Method} & \multicolumn{3}{c}{DIODE Outdoor}                & \multicolumn{3}{c}{ETH3D}                 & \multicolumn{3}{c}{Average}              \\\cmidrule(l){2-4} \cmidrule(l){5-7} \cmidrule(l){8-10} 
			& $\text{REL}^{p}\downarrow$   & $\delta_1^{p}\uparrow$    & Rk.$\downarrow$ & $\text{REL}^{p}\downarrow$    & $\delta_1^{p}\uparrow$    & Rk.$\downarrow$  & $\text{REL}^{p}\downarrow$   & $\delta_1^{p}\uparrow$    & Rk.$\downarrow$\\\midrule
			VGGT & 0.215 & 0.700 & 5.000 & 0.053 & 0.978 & 3.591 & 0.114 & 0.879 & 4.327 \\
			MoGe V2 & \colorbox{best2}{0.124} & \colorbox{best2}{0.841} & \colorbox{best2}{2.000} & \colorbox{best2}{0.044} & 0.980 & \colorbox{best2}{2.637} & \colorbox{best2}{0.064} & \colorbox{best2}{0.947} & \colorbox{best2}{2.191} \\
			WorldMirror & 0.155 & 0.788 & 3.045 & 0.049 & 0.976 & 3.023 & 0.086 & 0.914 & 3.032 \\
			MapAnything & 0.302 & 0.501 & 6.000 & 0.265 & 0.549 & 6.000 & 0.268 & 0.553 & 6.000 \\
			Pow3R & 0.197 & 0.750 & 3.955 & 0.074 & \colorbox{best2}{0.982} & 3.796 & 0.119 & 0.899 & 4.132 \\\midrule
			LDCM (Ours) & \colorbox{best}{0.077} & \colorbox{best}{0.949} & \colorbox{best}{1.000} & \colorbox{best}{0.039} & \colorbox{best}{0.994} & \colorbox{best}{1.728} & \colorbox{best}{0.037} & \colorbox{best}{0.983} & \colorbox{best}{1.164} \\
			\specialrule{1.5pt}{0.8ex}{0.8ex}
		\end{tabular}%
	}
	\label{tab:synthetic_affine_invariant_point}
	\vspace{-3mm}
\end{table}

\subsection{Ablation study}
\begin{table}[h]
	\caption{Ablation study on the coarse depth alignment strategy. We report the relative error (REL) for coarse depth and final prediction. The \colorbox{best}{best} and \colorbox{best2}{second-best} results are highlighted.}
	\resizebox{\columnwidth}{!}{%
		\begin{tabular}{ccccccccccc}
			\specialrule{1.5pt}{0.8ex}{0.8ex}
			\multicolumn{1}{c}{\multirow{2}{*}{Configuration}} & \multicolumn{5}{c}{Coarse Depth ($\text{REL}\downarrow$)}           & \multicolumn{5}{c}{Estimated Depth ($\text{REL}\downarrow$)}       \\\cmidrule{2-6}\cmidrule{7-11}
			\multicolumn{1}{c}{}                               & KITTI & iBims-1 & DIODE & ETH3D & Average & KITTI & iBims-1 & DIODE & ETH3D & Average \\\midrule
			Sparse                                            &   -    &     -    &   -    &   -    &     -    &     0.021  &     0.029    &  0.040    &   0.026    &  0.029       \\
			Global alignment                                   &    0.095   &   \colorbox{best2}{0.075}   &  \colorbox{best2}{0.102}   &   0.078   &  \colorbox{best2}{0.087}     &   \colorbox{best2}{0.020}     &      \colorbox{best2}{0.019}    &     \colorbox{best2}{0.035}   &    0.023   &    \colorbox{best2}{0.024}      \\
			LWLR                                               &   0.078    &    0.108     &   0.108    &    \colorbox{best2}{0.061}   &    0.088     &   \colorbox{best}{0.019}     &      0.022   &    0.036    &    \colorbox{best2}{0.021}    &      0.025   \\\midrule
			Poisson w/o global alignment &    \colorbox{best2}{0.069}   &   0.208     &   0.174    &  0.138  &   0.147  &   - &   -   &     -     &   -     &     -         \\
			Poisson                                           &    \colorbox{best}{0.033}   &    \colorbox{best}{0.073}     &   \colorbox{best}{0.088}    &    \colorbox{best}{0.044}   &    \colorbox{best}{0.059}     &   \colorbox{best}{0.019} &   \colorbox{best}{0.018}   &     \colorbox{best}{0.033}     &   \colorbox{best}{0.019}     &     \colorbox{best}{0.022}         \\
			\specialrule{1.5pt}{0.8ex}{0.8ex}
		\end{tabular}%
	}
	\label{tab:coarse_depth}
	\vspace{-5mm}
\end{table}
\begin{figure}[h]
	\includegraphics[width=\linewidth]{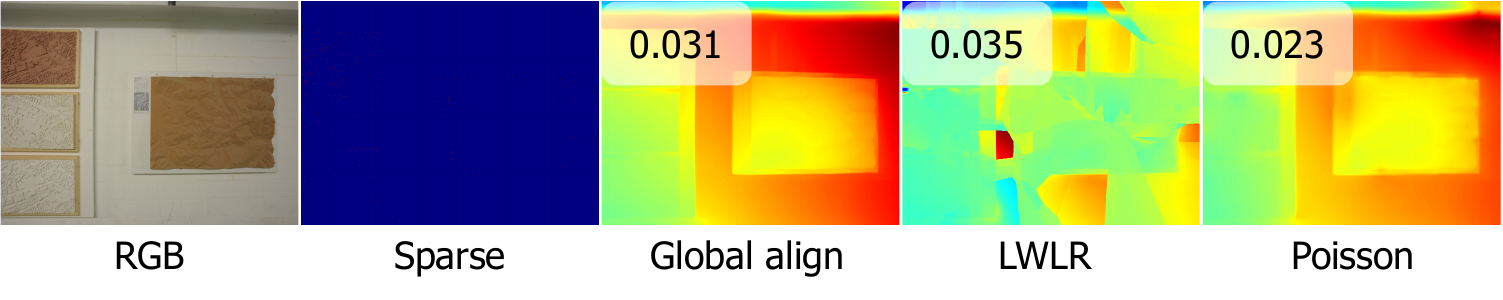}
	\vspace{-5mm}
	\caption{Qualitative comparison between three coarse alignment strategies. We report the relative error for each result.}
	\label{fig:ablation}
	
\end{figure}
We conduct ablation studies to evaluate the effectiveness of the Poisson-based coarse depth alignment strategy and the training objectives. For simplicity, we adopt LiDAR-simulated sparse patterns (64, 32, 16, and 8 lines) on outdoor datasets, and keypoint-based sampling on indoor datasets.

\textbf{Coarse Depth Alignment Strategy.} We ablate various coarse depth alignment strategies for robust geometric guidance. First, we assess the accuracy of the generated coarse depth maps. As shown on the left side of Table~\ref{tab:coarse_depth}, Poisson-based alignment achieves the best performance, demonstrating its effectiveness. Notably, global alignment is essential—its omission leads to a significant performance drop. By comparison, LWLR performs worse than even simple global alignment under extreme sparsity, highlighting its sensitivity to sparse and irregular inputs. A qualitative ablation example is provided in Fig.~\ref{fig:ablation}, where the Poisson-based method not only achieves the highest accuracy but also best preserves geometric structure. On the right side of Table~\ref{tab:coarse_depth}, we use these coarse depth maps as inputs to our completion model; again, the Poisson-based variant yields the best results.

\begin{table}[h]
	\caption{Ablation study on the output representation. We report the relative error (REL) for depth completion and $\text{REL}^{p}$ for point map estimation. The \colorbox{best}{best} and \colorbox{best2}{second-best} results are highlighted.}
	\resizebox{\columnwidth}{!}{%
		\begin{tabular}{ccccccccccc}
			\specialrule{1.5pt}{0.8ex}{0.8ex}
			\multirow{2}{*}{Configuration} & \multicolumn{5}{c}{Depth Completion ($\text{REL}\downarrow$)}      & \multicolumn{5}{c}{Point Map Estimation ($\text{REL}^{p}\downarrow$)}  \\\cmidrule{2-6}\cmidrule{7-11}
			& KITTI & iBims-1 & DIODE & ETH3D & Average & KITTI & iBims-1 & DIODE & ETH3D & Average \\\midrule
			SI-Log Depth                   &   0.023    &    0.023    &   \colorbox{best2}{ 0.037} &    \colorbox{best2}{0.021}   &   \colorbox{best2}{0.026}    &    -   &   -      &   -    &   -    &    -     \\
			SI-Log Depth + Ray map         &  \colorbox{best2}{0.022} & \colorbox{best2}{0.022} & 0.038 &  \colorbox{best2}{0.021} & \colorbox{best2}{0.026 }  &   \colorbox{best2}{0.073}   &    \colorbox{best2}{0.050}     &  \colorbox{best2}{0.084}    &   \colorbox{best2}{0.097}    &   \colorbox{best2}{0.067}   \\\midrule
			Point Map                      &    \colorbox{best}{0.019} &   \colorbox{best}{0.018}   &     \colorbox{best}{0.033}     &   \colorbox{best}{0.019}     &     \colorbox{best}{0.022}    &   \colorbox{best}{0.047}    &   \colorbox{best}{0.032}      &    \colorbox{best}{0.070}   &    \colorbox{best}{0.059}   &    \colorbox{best}{0.045}    \\
			\specialrule{1.5pt}{0.8ex}{0.8ex}
		\end{tabular}%
	}
	\label{tab:output}
	\vspace{-2mm}
\end{table}
\textbf{Output Representation.} We ablate the output representation by replacing the point map with either a conventional depth map or the concatenation of depth and dense ray maps (depth + ray map). As shown in Table~\ref{tab:output}, both alternatives lead to performance degradation, demonstrating that the point map provides more effective 3D structural guidance than depth-based representations.

\subsection{Depth completion results on standard benchmarks}

\begin{table}[h]
	\caption{\textbf{Quantitative comparison of depth completion methods on real-pattern benchmark datasets.} All methods are evaluated under zero-shot settings. Methods marked with $\dagger$ produce relative depth, and metric depth is recovered by optimizing global scale and shift via least squares regression using the same sparse depth prior. Methods marked with $\ddagger$ use dataset-specific configurations for indoor and outdoor scenes, respectively. The \colorbox{best}{best} and \colorbox{best2}{second-best} results are highlighted.}
	\resizebox{\columnwidth}{!}{%
		\begin{tabular}{cccccccccccccccc}
			\specialrule{1.5pt}{0.8ex}{0.8ex}
			\multirow{2}{*}{Method} & \multicolumn{5}{c}{NYUv2}                & \multicolumn{5}{c}{VOID}                 & \multicolumn{5}{c}{ETH3D}              \\\cmidrule(l){2-6} \cmidrule(l){7-11} \cmidrule(l){12-16} 
			& RMSE$\downarrow$ & MAE$\downarrow$   & REL$\downarrow$   & $\delta_1\uparrow$    & Rk.$\downarrow$ & RMSE$\downarrow$ & MAE$\downarrow$   & REL$\downarrow$   & $\delta_1\uparrow$  & Rk.$\downarrow$ & RMSE$\downarrow$ & MAE$\downarrow$   & REL$\downarrow$   & $\delta_1\uparrow$  & Rk.$\downarrow$ \\\midrule
			DepthPro & 0.332 & 0.253 & 0.096 & 0.929 & 15.000 & 0.759 & 0.396 & 0.189 & 0.726 & 14.833 & 3.199 & 2.562 & 0.302 & 0.477 & 14.750 \\
			UniDepth V1 & 0.213 & 0.148 & 0.056 & 0.981 & 10.375 & 0.651 & 0.267 & 0.107 & 0.902 & 12.083 & 3.482 & 3.170 & 0.579 & 0.116 & 15.625 \\
			UniDepth V2 & 0.293 & 0.218 & 0.085 & 0.948 & 14.000 & 0.651 & 0.269 & 0.115 & 0.900 & 13.000 & 1.630 & 1.169 & 0.200 & 0.726 & 13.000 \\
			DepthAnythingV2$\dagger$ & 0.220 & 0.128 & 0.045 & 0.977 & 11.250 & 0.605 & 0.214 & 0.063 & 0.958 & 8.250 & 1.915 & 0.493 & 0.063 & \colorbox{best2}{0.963} & 6.500 \\
			VGGT$\dagger$ & 0.168 & 0.087 & 0.033 & 0.985 & 7.000 & 0.572 & 0.196 & 0.064 & 0.952 & 6.750 & \colorbox{best2}{0.650} & 0.432 & 0.095 & 0.906 & 6.375 \\
			MoGe V1$\dagger$ & 0.180 & 0.093 & 0.037 & 0.979 & 8.750 & 0.577 & 0.200 & 0.064 & 0.952 & 7.500 & 2.877 & 0.450 & 0.108 & 0.924 & 6.500 \\
			MoGe V2 & 0.261 & 0.186 & 0.070 & 0.963 & 13.000 & 0.779 & 0.421 & 0.202 & 0.557 & 15.833 & 0.847 & 0.619 & 0.114 & 0.839 & 9.375 \\
			G2-MonoDepth$\ddagger$ & 0.166 & 0.071 & 0.026 & 0.985 & 7.125 & 0.607 & 0.195 & 0.055 & 0.942 & 7.500 & 1.425 & 0.525 & 0.136 & 0.886 & 10.375 \\
			OMNI-DC & 0.147 & 0.053 & 0.020 & 0.987 & 4.375 & 0.574 & \colorbox{best2}{0.168} & 0.040 & 0.962 & 4.167 & 0.822 & 0.317 & 0.079 & 0.925 & 4.625 \\
			PriorDA & \colorbox{best2}{0.122} & \colorbox{best2}{0.047} & \colorbox{best2}{0.017} & \colorbox{best2}{0.993} & 2.750 & 0.571 & 0.171 & \colorbox{best2}{0.039} & \colorbox{best2}{0.968} & \colorbox{best2}{3.333} & 0.671 & \colorbox{best2}{0.260} & \colorbox{best2}{0.061} & 0.962 & \colorbox{best2}{2.500} \\
			SPNet$\ddagger$ & 0.127 & \colorbox{best2}{0.047} & \colorbox{best2}{0.017} & 0.992 & \colorbox{best2}{2.500} & 0.578 & 0.178 & 0.054 & 0.959 & 5.250 & 1.299 & 0.372 & 0.092 & 0.943 & 6.625 \\
			PromptDA & 0.162 & 0.079 & 0.028 & 0.989 & 6.000 & \colorbox{best2}{0.565} & 0.182 & 0.049 & 0.965 & 4.000 & 0.911 & 0.483 & 0.090 & 0.896 & 6.875 \\
			WorldMirror$\dagger$ & 0.217 & 0.121 & 0.042 & 0.979 & 10.125 & 0.596 & 0.208 & 0.067 & 0.946 & 9.833 & 0.898 & 0.668 & 0.153 & 0.836 & 9.250 \\
			MapAnything & 0.724 & 0.327 & 0.132 & 0.885 & 16.000 & 0.782 & 0.282 & 0.110 & 0.900 & 13.750 & 2.283 & 0.874 & 0.150 & 0.863 & 12.375 \\
			Pow3R$\dagger$ & 0.155 & 0.081 & 0.031 & 0.988 & 5.625 & 0.571 & 0.196 & 0.067 & 0.949 & 7.333 & 0.881 & 0.656 & 0.154 & 0.833 & 9.875 \\\midrule
			LDCM (Ours) & \colorbox{best}{\textbf{0.113}} & \colorbox{best}{\textbf{0.037}} & \colorbox{best}{\textbf{0.013}} & \colorbox{best}{\textbf{0.994}} & \colorbox{best}{\textbf{1.000}} & \colorbox{best}{\textbf{0.536}} & \colorbox{best}{\textbf{0.145}} & \colorbox{best}{\textbf{0.028}} & \colorbox{best}{\textbf{0.977}} & \colorbox{best}{\textbf{1.000}} & \colorbox{best}{\textbf{0.445}} & \colorbox{best}{\textbf{0.154}} & \colorbox{best}{\textbf{0.035}} & \colorbox{best}{\textbf{0.978}} & \colorbox{best}{\textbf{1.250}} \\
			\specialrule{1.5pt}{0.8ex}{0.8ex}
		\end{tabular}%
	}
	\label{tab:benchmark_depth}
	\vspace{-4mm}
\end{table}

To further evaluate zero-shot depth completion under real-world sparse patterns, we follow prior work in evaluating methods on the NYUv2~\cite{nyuv2}, VOID~\cite{void}, and ETH3D~\cite{eth3d} datasets. For NYUv2, we adopt the sampling protocol from OMNI-DC~\cite{omnidc}, extracting 500 and 100 sparse depth points per image, respectively. For VOID, we use the provided sparse depth maps derived from a visual-inertial odometry system, which come in three sparsity levels: 1500, 500, and 150 points per frame. For ETH3D, we project the sparse 3D points from COLMAP SfM reconstructions into the image plane to generate sparse depth maps. Table~\ref{tab:benchmark_depth} reports the quantitative results on each dataset. As shown in the table, LDCM significantly outperforms all comparison methods, ranking first on all the datasets.

\section{Conclusion}
We have presented the Large Depth Completion Model (LDCM), a simple yet powerful framework for metric depth estimation from sparse observations. LDCM is both effective and robust, leveraging a Poisson-based alignment strategy to maximize the potential of existing monocular foundation models by preprocessing input sparse observations into strong geometric priors for subsequent feature learning. Furthermore, LDCM replaces the conventional depth map representation with a point map representation, enabling direct learning of the underlying 3D structure rather than per-pixel depth restoration. Our method achieves superior zero-shot performance across multiple benchmarks, demonstrating robustness under varying sparse observation patterns. Moreover, the point map design allows LDCM to naturally output metric-scaled 3D geometry without requiring camera intrinsics, facilitating reliable deployment in uncalibrated environments. We believe LDCM marks a significant advancement in depth completion and can serve as a robust foundational model for downstream 3D vision tasks.

\newpage
\section*{Acknowledgments}
This work was supported in part by the National Natural Science Foundation of China under grant 62301484, in part by the Jinhua Science and Technology Bureau Project under grant 2026-1-022, in part by the Young Talent Fund of Zhejiang Association for Science and Technology under grant ZJSKXQT2026135, and  in part by the Ningbo Natural Science Foundation of China under grant 2024J454.

\section*{Ethics Statement}
Our study focuses on depth completion, a core problem in the field of computer vision. The experimental evaluation is based exclusively on public datasets that have been curated without inclusion of any personally identifiable or sensitive content. We assert that this research has been carried out in accordance with the code of ethics.

\section*{Reproducibility Statement}
To facilitate verification and extension of our work, we include the implementation code in the supplementary materials. Furthermore, we provide key training and evaluation procedures in the paper, and will make the complete code and trained models publicly available after publication to support full experimental reproducibility.

\bibliography{iclr2026_conference}

@article{more,
	title={MoRE: 3D Visual Geometry Reconstruction Meets Mixture-of-Experts},
	author={Gao, Jingnan and Wang, Zhe and Fang, Xianze and Ren, Xingyu and Chen, Zhuo and Liu, Shengqi and Cheng, Yuhao and Lyu, Jiangjing and Yang, Xiaokang and Yan, Yichao},
	journal={arXiv preprint arXiv:2510.27234},
	year={2025}
}

@inproceedings{hypersim,
	title={Hypersim: A photorealistic synthetic dataset for holistic indoor scene understanding},
	author={Roberts, Mike and Ramapuram, Jason and Ranjan, Anurag and Kumar, Atulit and Bautista, Miguel Angel and Paczan, Nathan and Webb, Russ and Susskind, Joshua M},
	booktitle={Proceedings of the IEEE/CVF International Conference on Computer Vision},
	pages={10912--10922},
	year={2021}
}

@inproceedings{tartanair,
	title={Tartanair: A dataset to push the limits of visual slam},
	author={Wang, Wenshan and Zhu, Delong and Wang, Xiangwei and Hu, Yaoyu and Qiu, Yuheng and Wang, Chen and Hu, Yafei and Kapoor, Ashish and Scherer, Sebastian},
	booktitle={2020 IEEE/RSJ International Conference on Intelligent Robots and Systems},
	pages={4909--4916},
	year={2020},
}

@article{irs,
	title={Irs: A large naturalistic indoor robotics stereo dataset to train deep models for disparity and surface normal estimation},
	author={Wang, Qiang and Zheng, Shizhen and Yan, Qingsong and Deng, Fei and Zhao, Kaiyong and Chu, Xiaowen},
	journal={arXiv preprint arXiv:1912.09678},
	year={2019}
}

@inproceedings{pointodyssey,
	title={Pointodyssey: A large-scale synthetic dataset for long-term point tracking},
	author={Zheng, Yang and Harley, Adam W and Shen, Bokui and Wetzstein, Gordon and Guibas, Leonidas J},
	booktitle={Proceedings of the IEEE/CVF International Conference on Computer Vision},
	pages={19855--19865},
	year={2023}
}

@article{urbansyn,
	title={All for one, and one for all: Urbansyn dataset, the third musketeer of synthetic driving scenes},
	author={G{\'o}mez, Jose L and Silva, Manuel and Seoane, Antonio and Borr{\'a}s, Agn{\`e}s and Noriega, Mario and Ros, Germ{\'a}n and Iglesias-Guitian, Jose A and L{\'o}pez, Antonio M},
	journal={Neurocomputing},
	volume={637},
	pages={130038},
	year={2025},
}

@article{synscapes,
	title={Synscapes: A photorealistic synthetic dataset for street scene parsing},
	author={Wrenninge, Magnus and Unger, Jonas},
	journal={arXiv preprint arXiv:1810.08705},
	year={2018}
}

@inproceedings{matrixcity,
	title={Matrixcity: A large-scale city dataset for city-scale neural rendering and beyond},
	author={Li, Yixuan and Jiang, Lihan and Xu, Linning and Xiangli, Yuanbo and Wang, Zhenzhi and Lin, Dahua and Dai, Bo},
	booktitle={Proceedings of the IEEE/CVF International Conference on Computer Vision},
	pages={3205--3215},
	year={2023}
}

@misc{lightwheelocc,
	title={LightwheelOcc: A 3D Occupancy Synthetic Dataset in Autonomous Driving} ,
	author={LightwheelAI and LightwheelOcc contributors},
	howpublished={\url{https://github.com/OpenDriveLab/LightwheelOcc}},
	year={2024}
}

@inproceedings{mvssynth,
	title={Deepmvs: Learning multi-view stereopsis},
	author={Huang, Po-Han and Matzen, Kevin and Kopf, Johannes and Ahuja, Narendra and Huang, Jia-Bin},
	booktitle={Proceedings of the IEEE/CVF Conference on Computer Vision and Pattern Recognition},
	pages={2821--2830},
	year={2018}
}

@inproceedings{synthia,
	title={The synthia dataset: A large collection of synthetic images for semantic segmentation of urban scenes},
	author={Ros, German and Sellart, Laura and Materzynska, Joanna and Vazquez, David and Lopez, Antonio M},
	booktitle={Proceedings of the IEEE/CVF Conference on Computer Vision and Pattern Recognition},
	pages={3234--3243},
	year={2016}
}

@inproceedings{scannetpp,
	title={Scannet++: A high-fidelity dataset of 3d indoor scenes},
	author={Yeshwanth, Chandan and Liu, Yueh-Cheng and Nie{\ss}ner, Matthias and Dai, Angela},
	booktitle={Proceedings of the IEEE/CVF International Conference on Computer Vision},
	pages={12--22},
	year={2023}
}

@inproceedings{
	depthpro,
	title={Depth Pro: Sharp Monocular Metric Depth in Less Than a Second},
	author={Alexey Bochkovskiy and Ama{\"e}l Delaunoy and Hugo Germain and Marcel Santos and Yichao Zhou and Stephan Richter and Vladlen Koltun},
	booktitle={International Conference on Learning Representations},
	year={2025},
}

@inproceedings{unidepth,
	title={UniDepth: Universal monocular metric depth estimation},
	author={Piccinelli, Luigi and Yang, Yung-Hsu and Sakaridis, Christos and Segu, Mattia and Li, Siyuan and Van Gool, Luc and Yu, Fisher},
	booktitle={Proceedings of the IEEE/CVF Conference on Computer Vision and Pattern Recognition},
	pages={10106--10116},
	year={2024}
}

@article{unidepthv2,
	title={Unidepthv2: Universal monocular metric depth estimation made simpler},
	author={Piccinelli, Luigi and Sakaridis, Christos and Yang, Yung-Hsu and Segu, Mattia and Li, Siyuan and Abbeloos, Wim and Van Gool, Luc},
	journal={arXiv preprint arXiv:2502.20110},
	year={2025}
}

@article{depthanythingv2,
	title={Depth anything v2},
	author={Yang, Lihe and Kang, Bingyi and Huang, Zilong and Zhao, Zhen and Xu, Xiaogang and Feng, Jiashi and Zhao, Hengshuang},
	journal={Advances in Neural Information Processing Systems},
	volume={37},
	pages={21875--21911},
	year={2024}
}

@inproceedings{vggt,
	title={Vggt: Visual geometry grounded transformer},
	author={Wang, Jianyuan and Chen, Minghao and Karaev, Nikita and Vedaldi, Andrea and Rupprecht, Christian and Novotny, David},
	booktitle={Proceedings of the Computer Vision and Pattern Recognition Conference},
	pages={5294--5306},
	year={2025}
}

@inproceedings{moge,
	title={Moge: Unlocking accurate monocular geometry estimation for open-domain images with optimal training supervision},
	author={Wang, Ruicheng and Xu, Sicheng and Dai, Cassie and Xiang, Jianfeng and Deng, Yu and Tong, Xin and Yang, Jiaolong},
	booktitle={Proceedings of the Computer Vision and Pattern Recognition Conference},
	pages={5261--5271},
	year={2025}
}

@article{moge2,
	title={MoGe-2: Accurate Monocular Geometry with Metric Scale and Sharp Details},
	author={Wang, Ruicheng and Xu, Sicheng and Dong, Yue and Deng, Yu and Xiang, Jianfeng and Lv, Zelong and Sun, Guangzhong and Tong, Xin and Yang, Jiaolong},
	journal={arXiv preprint arXiv:2507.02546},
	year={2025}
}

@article{g2md,
	title={G2-monodepth: A general framework of generalized depth inference from monocular rgb+ x data},
	author={Wang, Haotian and Yang, Meng and Zheng, Nanning},
	journal={IEEE Transactions on Pattern Analysis and Machine Intelligence},
	volume={46},
	number={5},
	pages={3753--3771},
	year={2023},
}

@article{omnidc,
	title={OMNI-DC: Highly Robust Depth Completion with Multiresolution Depth Integration},
	author={Zuo, Yiming and Yang, Willow and Ma, Zeyu and Deng, Jia},
	journal={arXiv preprint arXiv:2411.19278},
	year={2024}
}

@article{spnet,
	title={Scale propagation network for generalizable depth completion},
	author={Wang, Haotian and Yang, Meng and Zheng, Xinhu and Hua, Gang},
	journal={IEEE Transactions on Pattern Analysis and Machine Intelligence},
	year={2024},
	publisher={IEEE}
}

@article{priorda,
	title={Depth Anything with Any Prior},
	author={Wang, Zehan and Chen, Siyu and Yang, Lihe and Wang, Jialei and Zhang, Ziang and Zhao, Hengshuang and Zhao, Zhou},
	journal={arXiv preprint arXiv:2505.10565},
	year={2025}
}

@article{marigolddc,
	title={Marigold-dc: Zero-shot monocular depth completion with guided diffusion},
	author={Viola, Massimiliano and Qu, Kevin and Metzger, Nando and Ke, Bingxin and Becker, Alexander and Schindler, Konrad and Obukhov, Anton},
	journal={arXiv preprint arXiv:2412.13389},
	year={2024}
}

@article{depthlab,
	title={Depthlab: From partial to complete},
	author={Liu, Zhiheng and Cheng, Ka Leong and Wang, Qiuyu and Wang, Shuzhe and Ouyang, Hao and Tan, Bin and Zhu, Kai and Shen, Yujun and Chen, Qifeng and Luo, Ping},
	journal={arXiv preprint arXiv:2412.18153},
	year={2024}
}

@inproceedings{nyuv2,
	title={Indoor segmentation and support inference from rgbd images},
	author={Silberman, Nathan and Hoiem, Derek and Kohli, Pushmeet and Fergus, Rob},
	booktitle={Proceedings of the European Conference on Computer Vision},
	pages={746--760},
	year={2012}
}

@inproceedings{marigold,
	title={Repurposing diffusion-based image generators for monocular depth estimation},
	author={Ke, Bingxin and Obukhov, Anton and Huang, Shengyu and Metzger, Nando and Daudt, Rodrigo Caye and Schindler, Konrad},
	booktitle={Proceedings of the IEEE/CVF Conference on Computer Vision and Pattern Recognition},
	pages={9492--9502},
	year={2024}
}

@inproceedings{kitti,
	title={Are we ready for autonomous driving? the kitti vision benchmark suite},
	author={Geiger, Andreas and Lenz, Philip and Urtasun, Raquel},
	booktitle={Proceedings of the IEEE/CVF Conference on Computer Vision and Pattern Recognition},
	pages={3354--3361},
	year={2012},
}

@inproceedings{kittidc,
	title={Sparsity invariant cnns},
	author={Uhrig, Jonas and Schneider, Nick and Schneider, Lukas and Franke, Uwe and Brox, Thomas and Geiger, Andreas},
	booktitle={International Conference on 3D Vision},
	pages={11--20},
	year={2017},
}

@article{diode,
	title={Diode: A dense indoor and outdoor depth dataset},
	author={Vasiljevic, Igor and Kolkin, Nick and Zhang, Shanyi and Luo, Ruotian and Wang, Haochen and Dai, Falcon Z and Daniele, Andrea F and Mostajabi, Mohammadreza and Basart, Steven and Walter, Matthew R and others},
	journal={arXiv preprint arXiv:1908.00463},
	year={2019}
}

@inproceedings{ibims,
	title={Evaluation of cnn-based single-image depth estimation methods},
	author={Koch, Tobias and Liebel, Lukas and Fraundorfer, Friedrich and Korner, Marco},
	booktitle={Proceedings of the European Conference on Computer Vision Workshops},
	pages={0--0},
	year={2018}
}

@article{void,
	title={Unsupervised Depth Completion From Visual Inertial Odometry},
	author={Wong, Alex and Fei, Xiaohan and Tsuei, Stephanie and Soatto, Stefano},
	journal={IEEE Robotics and Automation Letters},
	volume={5},
	number={2},
	pages={1899--1906},
	year={2020},
	publisher={IEEE}
}

@inproceedings{tpvd,
	title={Tri-perspective view decomposition for geometry-aware depth completion},
	author={Yan, Zhiqiang and Lin, Yuankai and Wang, Kun and Zheng, Yupeng and Wang, Yufei and Zhang, Zhenyu and Li, Jun and Yang, Jian},
	booktitle={Proceedings of the IEEE/CVF Conference on Computer Vision and Pattern Recognition},
	pages={4874--4884},
	year={2024}
}

@inproceedings{eth3d,
	title={A multi-view stereo benchmark with high-resolution images and multi-camera videos},
	author={Schops, Thomas and Schonberger, Johannes L and Galliani, Silvano and Sattler, Torsten and Schindler, Konrad and Pollefeys, Marc and Geiger, Andreas},
	booktitle={Proceedings of the IEEE/CVF Conference on Computer Vision and Pattern Recognition},
	pages={3260--3269},
	year={2017}
}

@inproceedings{promptda,
	title={Prompting depth anything for 4k resolution accurate metric depth estimation},
	author={Lin, Haotong and Peng, Sida and Chen, Jingxiao and Peng, Songyou and Sun, Jiaming and Liu, Minghuan and Bao, Hujun and Feng, Jiashi and Zhou, Xiaowei and Kang, Bingyi},
	booktitle={Proceedings of the Computer Vision and Pattern Recognition Conference},
	pages={17070--17080},
	year={2025}
}

@inproceedings{vit,
	title={An Image is Worth 16x16 Words: Transformers for Image Recognition at Scale},
	author={Dosovitskiy, Alexey and Beyer, Lucas and Kolesnikov, Alexander and Weissenborn, Dirk and Zhai, Xiaohua and Unterthiner, Thomas and Dehghani, Mostafa and Minderer, Matthias and Heigold, G and Gelly, S and others},
	booktitle={International Conference on Learning Representations},
	year={2020}
}

@article{dinov2,
	title={Dinov2: Learning robust visual features without supervision},
	author={Oquab, Maxime and Darcet, Timoth{\'e}e and Moutakanni, Th{\'e}o and Vo, Huy and Szafraniec, Marc and Khalidov, Vasil and Fernandez, Pierre and Haziza, Daniel and Massa, Francisco and El-Nouby, Alaaeldin and others},
	journal={arXiv preprint arXiv:2304.07193},
	year={2023}
}

@article{adamw,
	title={Decoupled weight decay regularization},
	author={Loshchilov, Ilya and Hutter, Frank},
	journal={arXiv preprint arXiv:1711.05101},
	year={2017}
}

@inproceedings{metric3d,
	title={Metric3d: Towards zero-shot metric 3d prediction from a single image},
	author={Yin, Wei and Zhang, Chi and Chen, Hao and Cai, Zhipeng and Yu, Gang and Wang, Kaixuan and Chen, Xiaozhi and Shen, Chunhua},
	booktitle={Proceedings of the IEEE/CVF International Conference on Computer Vision},
	pages={9043--9053},
	year={2023}
}

@article{metric3dv2,
	title={Metric3d v2: A versatile monocular geometric foundation model for zero-shot metric depth and surface normal estimation},
	author={Hu, Mu and Yin, Wei and Zhang, Chi and Cai, Zhipeng and Long, Xiaoxiao and Chen, Hao and Wang, Kaixuan and Yu, Gang and Shen, Chunhua and Shen, Shaojie},
	journal={IEEE Transactions on Pattern Analysis and Machine Intelligence},
	year={2024}
}

@inproceedings{lrru,
	title={Lrru: Long-short range recurrent updating networks for depth completion},
	author={Wang, Yufei and Li, Bo and Zhang, Ge and Liu, Qi and Gao, Tao and Dai, Yuchao},
	booktitle={Proceedings of the IEEE/CVF International Conference on Computer Vision},
	pages={9422--9432},
	year={2023}
}

@article{midas,
	title={Towards robust monocular depth estimation: Mixing datasets for zero-shot cross-dataset transfer},
	author={Ranftl, Ren{\'e} and Lasinger, Katrin and Hafner, David and Schindler, Konrad and Koltun, Vladlen},
	journal={IEEE Transactions on Pattern Analysis and Machine Intelligence},
	volume={44},
	number={3},
	pages={1623--1637},
	year={2020},
}

@inproceedings{dpt,
	title={Vision transformers for dense prediction},
	author={Ranftl, Ren{\'e} and Bochkovskiy, Alexey and Koltun, Vladlen},
	booktitle={Proceedings of the IEEE/CVF International Conference on Computer Vision},
	pages={12179--12188},
	year={2021}
}

@inproceedings{depthanything,
	title={Depth anything: Unleashing the power of large-scale unlabeled data},
	author={Yang, Lihe and Kang, Bingyi and Huang, Zilong and Xu, Xiaogang and Feng, Jiashi and Zhao, Hengshuang},
	booktitle={Proceedings of the IEEE/CVF conference on computer vision and pattern recognition},
	pages={10371--10381},
	year={2024}
}

@inproceedings{leres,
	title={Learning to recover 3d scene shape from a single image},
	author={Yin, Wei and Zhang, Jianming and Wang, Oliver and Niklaus, Simon and Mai, Long and Chen, Simon and Shen, Chunhua},
	booktitle={Proceedings of the IEEE/CVF conference on computer vision and pattern recognition},
	pages={204--213},
	year={2021}
}

@inproceedings{dust3r,
	title={Dust3r: Geometric 3d vision made easy},
	author={Wang, Shuzhe and Leroy, Vincent and Cabon, Yohann and Chidlovskii, Boris and Revaud, Jerome},
	booktitle={Proceedings of the IEEE/CVF Conference on Computer Vision and Pattern Recognition},
	pages={20697--20709},
	year={2024}
}

@inproceedings{mast3r,
	title={Grounding image matching in 3d with mast3r},
	author={Leroy, Vincent and Cabon, Yohann and Revaud, J{\'e}r{\^o}me},
	booktitle={Proceedings of the European Conference on Computer Vision},
	pages={71--91},
	year={2024}
}

@article{dens3r,
	title={Dens3R: A Foundation Model for 3D Geometry Prediction},
	author={Fang, Xianze and Gao, Jingnan and Wang, Zhe and Chen, Zhuo and Ren, Xingyu and Lyu, Jiangjing and Ren, Qiaomu and Yang, Zhonglei and Yang, Xiaokang and Yan, Yichao and others},
	journal={arXiv preprint arXiv:2507.16290},
	year={2025}
}

@inproceedings{rignet,
	title={RigNet: Repetitive image guided network for depth completion},
	author={Yan, Zhiqiang and Wang, Kun and Li, Xiang and Zhang, Zhenyu and Li, Jun and Yang, Jian},
	booktitle={Proceedings of the European Conference on Computer Vision},
	pages={214--230},
	year={2022},
	organization={Springer}
}

@inproceedings{pointdc,
	title={Aggregating feature point cloud for depth completion},
	author={Yu, Zhu and Sheng, Zehua and Zhou, Zili and Luo, Lun and Cao, Si-Yuan and Gu, Hong and Zhang, Huaqi and Shen, Hui-Liang},
	booktitle={Proceedings of the IEEE/CVF International Conference on Computer Vision},
	pages={8732--8743},
	year={2023}
}

@article{tpvd2,
	title={Tri-Perspective View Decomposition for Geometry Aware Depth Completion and Super-Resolution},
	author={Yan, Zhiqiang and Wang, Kun and Li, Xiang and Gao, Guangwei and Li, Jun and Yang, Jian},
	journal={IEEE Transactions on Pattern Analysis and Machine Intelligence},
	year={2025}
}

@article{rignetpp,
	title={RigNet++: Semantic Assisted Repetitive Image Guided Network for Depth Completion: Z. Yan et al.},
	author={Yan, Zhiqiang and Li, Xiang and Hui, Le and Zhang, Zhenyu and Li, Jun and Yang, Jian},
	journal={International Journal of Computer Vision},
	pages={1--23},
	year={2025}
}

@article{spn,
	title={Learning affinity via spatial propagation networks},
	author={Liu, Sifei and De Mello, Shalini and Gu, Jinwei and Zhong, Guangyu and Yang, Ming-Hsuan and Kautz, Jan},
	journal={Advances in Neural Information Processing Systems},
	volume={30},
	year={2017}
}

@inproceedings{cspn,
	title={Depth estimation via affinity learned with convolutional spatial propagation network},
	author={Cheng, Xinjing and Wang, Peng and Yang, Ruigang},
	booktitle={Proceedings of the European Conference on Computer Vision},
	pages={103--119},
	year={2018}
}

@article{cspn2,
	title={Learning depth with convolutional spatial propagation network},
	author={Cheng, Xinjing and Wang, Peng and Yang, Ruigang},
	journal={IEEE Transactions on Pattern Analysis and Machine Intelligence},
	volume={42},
	number={10},
	pages={2361--2379},
	year={2019}
}

@inproceedings{nlspn,
	title={Non-local spatial propagation network for depth completion},
	author={Park, Jinsun and Joo, Kyungdon and Hu, Zhe and Liu, Chi-Kuei and So Kweon, In},
	booktitle={Proceedings of the European conference on computer vision},
	pages={120--136},
	year={2020}
}

@inproceedings{dyspn,
	title={Dynamic spatial propagation network for depth completion},
	author={Lin, Yuankai and Cheng, Tao and Zhong, Qi and Zhou, Wending and Yang, Hua},
	booktitle={Proceedings of the AAAI Conference on Artificial Intelligence},
	volume={36},
	pages={1638--1646},
	year={2022}
}

@article{pacgdc,
	title={PacGDC: Label-Efficient Generalizable Depth Completion with Projection Ambiguity and Consistency},
	author={Wang, Haotian and Xiao, Aoran and Zhang, Xiaoqin and Yang, Meng and Lu, Shijian},
	journal={arXiv preprint arXiv:2507.07374},
	year={2025}
}

@inproceedings{pow3r,
	title={Pow3r: Empowering unconstrained 3d reconstruction with camera and scene priors},
	author={Jang, Wonbong and Weinzaepfel, Philippe and Leroy, Vincent and Agapito, Lourdes and Revaud, Jerome},
	booktitle={Proceedings of the Computer Vision and Pattern Recognition Conference},
	pages={1071--1081},
	year={2025}
}

@inproceedings{depthprompting,
	title={Depth prompting for sensor-agnostic depth estimation},
	author={Park, Jin-Hwi and Jeong, Chanhwi and Lee, Junoh and Jeon, Hae-Gon},
	booktitle={Proceedings of the IEEE/CVF Conference on Computer Vision and Pattern Recognition},
	pages={9859--9869},
	year={2024}
}

@article{gpt4,
	title={Gpt-4 technical report},
	author={Achiam, Josh and Adler, Steven and Agarwal, Sandhini and Ahmad, Lama and Akkaya, Ilge and Aleman, Florencia Leoni and Almeida, Diogo and Altenschmidt, Janko and Altman, Sam and Anadkat, Shyamal and others},
	journal={arXiv preprint arXiv:2303.08774},
	year={2023}
}

@article{qwen3,
	title={Qwen3 technical report},
	author={Yang, An and Li, Anfeng and Yang, Baosong and Zhang, Beichen and Hui, Binyuan and Zheng, Bo and Yu, Bowen and Gao, Chang and Huang, Chengen and Lv, Chenxu and others},
	journal={arXiv preprint arXiv:2505.09388},
	year={2025}
}

@article{lwlr,
	title={Towards 3D Scene Reconstruction from Locally Scale-Aligned Monocular Video Depth},
	author={Xu, Guangkai and Yin, Wei and Chen, Hao and Shen, Chunhua and Cheng, Kai and Wu, Feng and Zhao, Feng},
	journal={arXiv preprint arXiv:2202.01470},
	year={2022}
}

@article{conjugate,
	title={Methods of conjugate gradients for solving linear systems},
	author={Hestenes, Magnus R and Stiefel, Eduard and others},
	journal={Journal of research of the National Bureau of Standards},
	volume={49},
	number={6},
	pages={409--436},
	year={1952}
}

@article{AR,
	title={Optimizing depth perception in virtual and augmented reality through gaze-contingent stereo rendering},
	author={Krajancich, Brooke and Kellnhofer, Petr and Wetzstein, Gordon},
	journal={ACM transactions on graphics},
	volume={39},
	number={6},
	pages={1--10},
	year={2020}
}

@inproceedings{embodiedscan,
	title={Embodiedscan: A holistic multi-modal 3d perception suite towards embodied ai},
	author={Wang, Tai and Mao, Xiaohan and Zhu, Chenming and Xu, Runsen and Lyu, Ruiyuan and Li, Peisen and Chen, Xiao and Zhang, Wenwei and Chen, Kai and Xue, Tianfan and others},
	booktitle={Proceedings of the IEEE/CVF Conference on Computer Vision and Pattern Recognition},
	pages={19757--19767},
	year={2024}
}

@article{bevbert,
	title={Bevbert: Multimodal map pre-training for language-guided navigation},
	author={An, Dong and Qi, Yuankai and Li, Yangguang and Huang, Yan and Wang, Liang and Tan, Tieniu and Shao, Jing},
	journal={arXiv preprint arXiv:2212.04385},
	year={2022}
}

@inproceedings{bevdc,
	title={Bev@dc: Bird's-eye view assisted training for depth completion},
	author={Zhou, Wending and Yan, Xu and Liao, Yinghong and Lin, Yuankai and Huang, Jin and Zhao, Gangming and Cui, Shuguang and Li, Zhen},
	booktitle={Proceedings of the IEEE/CVF Conference on Computer Vision and Pattern Recognition},
	pages={9233--9242},
	year={2023}
}

@inproceedings{voxformer,
	title={Voxformer: Sparse voxel transformer for camera-based 3d semantic scene completion},
	author={Li, Yiming and Yu, Zhiding and Choy, Christopher and Xiao, Chaowei and Alvarez, Jose M and Fidler, Sanja and Feng, Chen and Anandkumar, Anima},
	booktitle={Proceedings of the IEEE/CVF Conference on Computer Vision and Pattern Recognition},
	pages={9087--9098},
	year={2023}
}

@inproceedings{sam,
	title={Segment anything},
	author={Kirillov, Alexander and Mintun, Eric and Ravi, Nikhila and Mao, Hanzi and Rolland, Chloe and Gustafson, Laura and Xiao, Tete and Whitehead, Spencer and Berg, Alexander C and Lo, Wan-Yen and others},
	booktitle={Proceedings of the IEEE/CVF international conference on computer vision},
	pages={4015--4026},
	year={2023}
}

@inproceedings{foundationstereo,
	title={Foundationstereo: Zero-shot stereo matching},
	author={Wen, Bowen and Trepte, Matthew and Aribido, Joseph and Kautz, Jan and Gallo, Orazio and Birchfield, Stan},
	booktitle={Proceedings of the Computer Vision and Pattern Recognition Conference},
	pages={5249--5260},
	year={2025}
}

@inproceedings{defomstereo,
	title={Defom-stereo: Depth foundation model based stereo matching},
	author={Jiang, Hualie and Lou, Zhiqiang and Ding, Laiyan and Xu, Rui and Tan, Minglang and Jiang, Wenjie and Huang, Rui},
	booktitle={Proceedings of the Computer Vision and Pattern Recognition Conference},
	pages={21857--21867},
	year={2025}
}

@inproceedings{monster,
	title={Monster: Marry monodepth to stereo unleashes power},
	author={Cheng, Junda and Liu, Longliang and Xu, Gangwei and Wang, Xianqi and Zhang, Zhaoxing and Deng, Yong and Zang, Jinliang and Chen, Yurui and Cai, Zhipeng and Yang, Xin},
	booktitle={Proceedings of the Computer Vision and Pattern Recognition Conference},
	pages={6273--6282},
	year={2025}
}

@article{ducos,
	title={DuCos: Duality Constrained Depth Super-Resolution via Foundation Model},
	author={Yan, Zhiqiang and Wang, Zhengxue and Dong, Haoye and Li, Jun and Yang, Jian and Lee, Gim Hee},
	journal={arXiv preprint arXiv:2503.04171},
	year={2025}
}

@article{worldmirror,
	title={WorldMirror: Universal 3D World Reconstruction with Any-Prior Prompting},
	author={Liu, Yifan and Min, Zhiyuan and Wang, Zhenwei and Wu, Junta and Wang, Tengfei and Yuan, Yixuan and Luo, Yawei and Guo, Chunchao},
	journal={arXiv preprint arXiv:2510.10726},
	year={2025}
}

@misc{mapanything,
	title={{MapAnything}: Universal Feed-Forward Metric {3D} Reconstruction},
	author={Nikhil Keetha and Norman M\"{u}ller and Johannes Sch\"{o}nberger and Lorenzo Porzi and Yuchen Zhang and Tobias Fischer and Arno Knapitsch and Duncan Zauss and Ethan Weber and Nelson Antunes and Jonathon Luiten and Manuel Lopez-Antequera and Samuel Rota Bul\`{o} and Christian Richardt and Deva Ramanan and Sebastian Scherer and Peter Kontschieder},
	note={arXiv preprint arXiv:2509.13414},
	year={2025}
}

@article{anysplat,
	title={AnySplat: Feed-forward 3D Gaussian Splatting from Unconstrained Views},
	author={Jiang, Lihan and Mao, Yucheng and Xu, Linning and Lu, Tao and Ren, Kerui and Jin, Yichen and Xu, Xudong and Yu, Mulin and Pang, Jiangmiao and Zhao, Feng and others},
	journal={arXiv preprint arXiv:2505.23716},
	year={2025}
}

@inproceedings{testpromptdc,
	title={Test-Time Prompt Tuning for Zero-Shot Depth Completion},
	author={Jeong, Chanhwi and Bae, Inhwan and Park, Jin-Hwi and Jeon, Hae-Gon},
	booktitle={Proceedings of the IEEE/CVF International Conference on Computer Vision},
	pages={9443--9454},
	year={2025}
}

@inproceedings{drivingstereo,
	title={Drivingstereo: A large-scale dataset for stereo matching in autonomous driving scenarios},
	author={Yang, Guorun and Song, Xiao and Huang, Chaoqin and Deng, Zhidong and Shi, Jianping and Zhou, Bolei},
	booktitle={Proceedings of the IEEE/CVF Conference on Computer Vision and Pattern Recognition},
	pages={899--908},
	year={2019}
}

@article{fe2e,
	title={From editor to dense geometry estimator},
	author={Wang, JiYuan and Lin, Chunyu and Sun, Lei and Liu, Rongying and Nie, Lang and Li, Mingxing and Liao, Kang and Chu, Xiangxiang and Zhao, Yao},
	journal={arXiv preprint arXiv:2509.04338},
	year={2025}
}

@article{jasmine,
	title={Jasmine: Harnessing diffusion prior for self-supervised depth estimation},
	author={Wang, Jiyuan and Lin, Chunyu and Guan, Cheng and Nie, Lang and He, Jing and Li, Haodong and Liao, Kang and Zhao, Yao},
	journal={arXiv preprint arXiv:2503.15905},
	year={2025}
}

@article{cgformer,
	title={Context and geometry aware voxel transformer for semantic scene completion},
	author={Yu, Zhu and Zhang, Runmin and Ying, Jiacheng and Yu, Junchen and Hu, Xiaohai and Luo, Lun and Cao, Si-Yuan and Shen, Hui-Liang},
	journal={Advances in Neural Information Processing Systems},
	volume={37},
	pages={1531--1555},
	year={2024}
}

@article{voxdet,
	title={Voxdet: Rethinking 3d semantic occupancy prediction as dense object detection},
	author={Li, Wuyang and Yu, Zhu and Alahi, Alexandre},
	journal={Advances in Neural Information Processing Systems},
	volume={38},
	year={2025},
	publisher={Neural Information Processing Systems Foundation, Inc.(NeurIPS)}
}

@inproceedings{locc,
	title={Language driven occupancy prediction},
	author={Yu, Zhu and Pang, Bowen and Liu, Lizhe and Zhang, Runmin and Li, Qiang and Cao, Si-Yuan and Luo, Maochun and Chen, Mingxia and Yang, Sheng and Shen, Hui-Liang},
	booktitle={Proceedings of the IEEE/CVF International Conference on Computer Vision},
	pages={7548--7558},
	year={2025}
}
\bibliographystyle{iclr2026_conference}
\newpage
\section*{Appendix}
\appendix

\section{Datasets}
\subsection{Training Datasets}
We collected 11 open-source RGB-D datasets to train LDCM, comprising 10 synthetic and 1 real-world dataset. An overview of the training datasets is provided in Table~\ref{tab:training_datasets}, spanning four distinct domains: indoor, outdoor, in-the-wild, and driving scenarios. The combined training set contains approximately 2.6 million images. The number of RGB-D pairs in each dataset may slightly differ from the originally released versions, as we manually excluded some invalid frames.
\begin{table}[h]
	\centering
	\caption{An overview of the training datasets.}
	\resizebox{\linewidth}{!}{
		\begin{tabular}{cccc}
			\toprule
			Dataset & Domain & Statistic & Type \\
			\midrule
			Hypersim~\cite{hypersim}	 & Indoor & 75K & Synthetic \\
			TartanAir~\cite{tartanair} & In-the-wild & 306K & Synthetic \\
			IRS~\cite{irs} & Indoor & 101K & Synthetic \\
			PointOdyssey~\cite{pointodyssey} & Indoor & 303K & Synthetic \\
			UrbanSyn~\cite{urbansyn} & Outdoor/Driving & 7K & Synthetic \\
			Synscapes~\cite{synscapes} & Outdoor/Driving & 25K & Synthetic \\
			MatrixCity~\cite{matrixcity} & Outdoor/Driving & 424K & Synthetic \\
			LightwheelOcc~\cite{lightwheelocc} & Outdoor/Driving & 204K & Synthetic \\
			MVS-Synth~\cite{mvssynth} & Outdoor/Driving & 12K & Synthetic \\
			Synthia~\cite{synthia} & Outdoor/Driving & 140K & Synthetic \\
			ScanNet++~\cite{scannetpp} & Indoor & 1M & Real \\\midrule
			Total & - & 2.6M & -\\
			\bottomrule
		\end{tabular}
	}
	\label{tab:training_datasets}
\end{table}

\subsection{Evaluation Datasets}
We use six datasets that are excluded from the training set to compare the performance between LDCM and previous state-of-the-art methods. Below, we provide details for each dataset.

\textbf{NYUv2 Dataset.} The NYUv2 dataset~\cite{nyuv2} is an indoor dataset captured using a Microsoft Kinect sensor, containing RGB and depth sequences from 464 indoor scenes. The official test split contains 654 samples. Following Marigold~\cite{marigold}, we crop the images to a resolution of $\text{426} \times \text{560}$ for consistent input dimensions.

\textbf{KITTI Dataset.} The KITTI Depth dataset~\cite{kitti,kittidc} is a large-scale outdoor dataset collected from a moving vehicle. The official validation split consists of 1,000 samples. Depth maps are acquired using an HDL-64 LiDAR sensor, with raw depth maps containing fewer than $\text{6}\%$ valid pixels. The provided ground truth is generated by fusing multiple consecutive LiDAR scans, resulting in a denser depth map with approximately $\text{14}\%$ valid pixels. For depth completion, input images are center-cropped to the bottom region of $\text{252} \times \text{1216}$ to exclude the sky and regions with unreliable depth due to the limited vertical field of view of the LiDAR.   

\textbf{DIODE Dataset.}  The DIODE dataset~\cite{diode} contains thousands of high-resolution RGB images with accurate, dense, and long-range depth measurements, captured using a FARO Focus S350 laser scanner. The official validation split includes 3 indoor and 3 outdoor scenes, comprising 325 and 446 samples, respectively. To reduce noise at occlusion boundaries, we filter out depth values where the maximum relative difference to any neighboring pixel exceeds $\text{5}\%$ (indoor) and $\text{15}\%$ (outdoor). Input images are resized to $\text{480} \times \text{640}$.

\textbf{iBims-1 Dataset.} The iBims-1 dataset~\cite{ibims} is an indoor benchmark captured in diverse environments, providing high-resolution RGB images and highly accurate depth maps derived from laser scans. The official evaluation split contains 100 samples, with images at a native resolution of $\text{480} \times \text{640}$.

\textbf{VOID Dataset.} The VOID dataset~\cite{void} is an indoor dataset captured using the Intel RealSense D435i camera. The official validation split consists of 800 samples, each paired with sparse depth maps at three sparsity levels (approximately 1500, 500, and 150 valid pixels) and RGB images at a resolution of $\text{480} \times \text{640}$. These varying sparsity levels allow for robust evaluation under different input conditions.

\textbf{ETH3D Dataset.} The ETH3D dataset~\cite{eth3d} consists of multi-view stereo images and dense depth maps captured using a high-precision laser scanner and DSLR cameras, covering diverse viewpoints and scene types. The official validation set contains 13 scenes with a total of 454 image pairs. The original image resolution is $\text{4032} \times \text{6048}$. Input images are resized to $\text{480} \times \text{640}$.

\section{Evaluation Details}
\subsection{Comparison Methods}
We compare LDCM against a comprehensive set of state-of-the-art approaches: 
DepthPro\footnote{\url{https://github.com/apple/ml-depth-pro}.}~\cite{depthpro}, 
UniDepth V1 \& V2\footnote{\url{https://github.com/lpiccinelli-eth/UniDepth}.}~\cite{unidepth, unidepthv2}, 
Depth Anything V2\footnote{\url{https://github.com/DepthAnything/Depth-Anything-V2}.}~\cite{depthanythingv2}, 
VGGT\footnote{\url{https://github.com/facebookresearch/vggt}.}~\cite{vggt}, 
MoGe V1 \& V2\footnote{\url{https://github.com/microsoft/MoGe}.}~\cite{moge, moge2}, 
G2-MonoDepth\footnote{\url{https://github.com/Wang-xjtu/G2-MonoDepth}.}~\cite{g2md}, 
OMNI-DC\footnote{\url{https://github.com/princeton-vl/OMNI-DC}.}~\cite{omnidc}, 
PriorDA\footnote{\url{https://github.com/SpatialVision/Prior-Depth-Anything}.}~\cite{priorda}, 
SPNet\footnote{\url{https://github.com/Wang-xjtu/SPNet}.}~\cite{spnet}, 
PromptDA\footnote{\url{https://github.com/DepthAnything/PromptDA}.}~\cite{promptda},
Marigold-DC\footnote{\url{https://github.com/prs-eth/Marigold-DC}.}~\cite{marigolddc},
DepthLab\footnote{\url{https://github.com/ant-research/DepthLab}.}~\cite{depthlab},
Pow3R\footnote{\url{https://github.com/naver/pow3r}.}~\cite{pow3r},
MapAnything\footnote{\url{https://github.com/facebookresearch/map-anything}.}~\cite{mapanything},
WorldMirror\footnote{\url{https://github.com/Tencent-Hunyuan/HunyuanWorld-Mirror}.}~\cite{worldmirror},
spanning the key tasks of monocular depth estimation, monocular geometry estimation, depth completion. All methods are evaluated using their publicly available implementations and pre-trained checkpoints. Notably, G2-MonoDepth~\cite{g2md} and SPNet~\cite{spnet} employ different configurations for indoor and outdoor scenarios, while LDCM and the remaining methods do not use scenario-specific hyperparameters. PromptDA~\cite{promptda} is specifically designed to leverage dense, low-resolution priors; therefore, we apply Poisson-based reconstruction to the input sparse depth map to obtain a dense prior before inference. Pow3R~\cite{pow3r} and WorldMirror~\cite{worldmirror} produce relative geometry, even when sparse depth priors are provided. 

\subsection{EVALUATION PROTOCOL}
To clarify the notation in this section:
\begin{itemize}
	\item $\mathbf{P}$ and  $\hat{\mathbf{P}}$ are the predicted and ground truth points, respectively.
	\item $\mathbf{D}$ and  $\hat{\mathbf{D}}$ are the predicted and ground truth depths, which are the z-coordinate of corresponding points.
	\item $\mathcal{M}$ is the mask of valid ground truth.
\end{itemize}

\textbf{Depth Completion.} In the manuscript, we use four standard metrics for depth completion evaluation, including RMSE, MAE, REL, $\delta_{1}$. Formally, they are defined as follows:
\begin{itemize}
	\item Root mean square error (RMSE):\begin{equation}\sqrt{\frac{1}{\left|\mathcal{M}\right|}\sum_{i\in\mathcal{M}}(\hat{\mathbf{D}}_{i}-\mathbf{D}_{i})^2}\end{equation}
	\item Mean absolute error (MAE):
	\begin{equation}\frac{1}{\left|\mathcal{M}\right|}\sum_{i\in\mathcal{M}}\left|\hat{\mathbf{D}}_{i}-\mathbf{D}_{i}\right|\end{equation}
	\item Mean relative error (REL):
	\begin{equation}\frac{1}{\left|\mathcal{M}\right|}\sum_{i\in\mathcal{M}}\frac{\left|\hat{\mathbf{D}}_{i}-\mathbf{D}_{i}\right|}{\hat{\mathbf{D}}_{i}}\end{equation}
	\item Thresholded accuracy ($\delta_{1}$):
	\begin{equation}
		\frac{1}{|\mathcal{M}|}
		\sum_{i \in \mathcal{M}}
		\left[
		\max\left(
		\frac{\hat{D}_i}{D_i},
		\frac{D_i}{\hat{D}_i}
		\right) < 1.25
		\right].	
	\end{equation}
\end{itemize}

For models that produce relative depth maps $\mathbf{D}_r$ , we first follow Equation~\ref{eq:global_align} to compute $(\alpha, \beta)$, and then the metric depth maps are recovered	 by:
\begin{equation}
	\mathbf{D} = \alpha\cdot\mathbf{D}_r + \beta.
\end{equation}

\textbf{Point Map Estimation.} For evaluating the reconstructed 3D point map, we adopt analogous metrics based on Euclidean distances between predicted and ground truth points. The metrics include $\mathrm{RMSE}^p$, $\mathrm{MAE}^p$, $\mathrm{REL}^p$, and $\delta_{1}^p$, defined as:
\begin{itemize}
	\item Point-wise Root Mean Square Error ($\mathrm{RMSE}^p$):
	\begin{equation}
		\sqrt{ \frac{1}{|\mathcal{M}|} \sum_{i \in \mathcal{M}} \left\| \hat{\mathbf{P}}_i - \mathbf{P}_i \right\|^2 }
	\end{equation}
	
	\item Point-wise Mean Absolute Error ($\mathrm{MAE}^p$):
	\begin{equation}
		\frac{1}{|\mathcal{M}|} \sum_{i \in \mathcal{M}} \left\| \hat{\mathbf{P}}_i - \mathbf{P}_i \right\|
	\end{equation}
	
	\item Point-wise Mean Relative Error ($\mathrm{REL}^p$):
	\begin{equation}
		\frac{1}{|\mathcal{M}|} \sum_{i \in \mathcal{M}} \frac{ \left\| \hat{\mathbf{P}}_i - \mathbf{P}_i \right\| }{ \left\| \hat{\mathbf{P}}_i \right\| }
	\end{equation}
	
	\item Point-wise Thresholded Accuracy ($\delta_{1}^p$):
	\begin{equation}
		\frac{1}{|\mathcal{M}|}
		\sum_{i \in \mathcal{M}}
		\left[
		\|\hat{P}_i - P_i\|
		<
		0.25 \cdot \min(\|P_i\|, \|\hat{P}_i\|)
		\right].	
	\end{equation}
\end{itemize}

\textbf{Affine-invariant Point Map Estimation.} To evaluate the affine-invariant point map, we first compute the scale $\alpha_{p}$ and shift $\beta_{p}$ using the following equation, which recovers the affine transformation applied to the predicted point map. This equation can be solved efficiently using the ROE solver proposed by MoGe~\cite{moge}.
\begin{equation}
	(\alpha_{p}, \beta_{p}) = \arg\min_{\alpha_{p}, \beta_{p}} \sum_{i \in \mathcal{M}} \left( \hat{\mathbf{P}}_{i} - \alpha_{p} \cdot \mathbf{P}_{i} - \beta_{p} \right)^2,\label{eq:affine_align}
\end{equation}

\section{More Comparison Results with Diffusion-based methods}
Here, we present additional comparisons with diffusion-based models—Marigold-DC~\cite{marigolddc} and DepthLab~\cite{depthlab}. Due to their prohibitively long inference times, we evaluate these methods primarily on three benchmarks with varying levels of sparse input: NYUv2~\cite{nyuv2} (500 and 100 points), VOID~\cite{void} (150, 500, and 1500 points), and KITTI~\cite{kitti} (64, 32, and 16 scan lines). As shown in Table~\ref{tab:depth_diffusion}, LDCM consistently outperforms both Marigold-DC and DepthLab across all settings.

\begin{table}[]
	\caption{\textbf{Quantitative comparison of depth completion with diffusion-based methods on benchmark datasets.} The \textbf{best} results are in \textbf{bold}.}
	\resizebox{\columnwidth}{!}{%
		\begin{tabular}{ccccccccccccc}
			\specialrule{1.5pt}{0.8ex}{0.8ex}
			\multirow{2}{*}{Method} & \multicolumn{4}{c}{VOID-1500-Points} & \multicolumn{4}{c}{VOID-500-Points}  & \multicolumn{4}{c}{VOID-150-Points}  \\\cmidrule{2-5}\cmidrule{6-9}\cmidrule{10-13}
			& RMSE$\downarrow$ & MAE$\downarrow$   & REL$\downarrow$   & $\delta_1\uparrow$   & RMSE$\downarrow$ & MAE$\downarrow$   & REL$\downarrow$   & $\delta_1\uparrow$ & RMSE$\downarrow$ & MAE$\downarrow$   & REL$\downarrow$   & $\delta_1\uparrow$      \\\midrule
			DepthLab                & 0.577 & 0.162 & 0.034 & 0.969 & 0.572 & 0.183 & 0.053 & 0.941 & 0.688 & 0.249 & 0.083 & 0.901 \\
			Marigold-DC             & 0.553 & 0.154 & 0.031 & 0.975 & 0.536 & 0.162 & 0.043 & 0.965 & 0.626 & 0.199 & 0.053 & 0.955 \\\midrule
			LDCM (Ours)                    & \textbf{0.528} & \textbf{0.135} & \textbf{0.021 }& \textbf{0.981} & \textbf{0.501} & \textbf{0.134} & \textbf{0.027} & \textbf{0.978} & \textbf{0.580} & \textbf{0.167} & \textbf{0.035} & \textbf{0.972} \\\specialrule{1pt}{0.4ex}{0.4ex}
			\multirow{2}{*}{Method} & \multicolumn{4}{c}{NYUv2-500-Points} & \multicolumn{4}{c}{NYUv2-100-Points} & \multicolumn{4}{c}{KITTI-64-Lines}  \\\cmidrule{2-5}\cmidrule{6-9}\cmidrule{10-13}
			& RMSE$\downarrow$ & MAE$\downarrow$   & REL$\downarrow$   & $\delta_1\uparrow$   & RMSE$\downarrow$ & MAE$\downarrow$   & REL$\downarrow$   & $\delta_1\uparrow$ & RMSE$\downarrow$ & MAE$\downarrow$   & REL$\downarrow$   & $\delta_1\uparrow$      \\\midrule
			DepthLab                & 0.118 & 0.041 & 0.015 & 0.993 & 0.213 & 0.100 & 0.037 & 0.976 & 2.032 & 0.828 & 0.061 & 0.962 \\
			Marigold-DC             & 0.116 & 0.040 & 0.014 & 0.993 & 0.157 & 0.061 & 0.022 & 0.988 & 1.931 & 0.818 & 0.054 & 0.971 \\\midrule
			LDCM (Ours)                      & \textbf{0.094} & \textbf{0.028} & \textbf{0.009} & \textbf{0.996} & \textbf{0.131} & \textbf{0.045} & \textbf{0.016} & \textbf{0.992} & \textbf{1.240} & \textbf{0.292} & \textbf{0.016} & \textbf{0.993} \\\specialrule{1pt}{0.4ex}{0.4ex}
			\multirow{2}{*}{Method} & \multicolumn{4}{c}{KITTI-32-Lines}  & \multicolumn{4}{c}{KITTI-16-Lines}  & \multicolumn{4}{c}{AVERAGE}   \\\cmidrule{2-5}\cmidrule{6-9}\cmidrule{10-13}
			& RMSE$\downarrow$ & MAE$\downarrow$   & REL$\downarrow$   & $\delta_1\uparrow$   & RMSE$\downarrow$ & MAE$\downarrow$   & REL$\downarrow$   & $\delta_1\uparrow$ & RMSE$\downarrow$ & MAE$\downarrow$   & REL$\downarrow$   & $\delta_1\uparrow$      \\\midrule
			DepthLab                & 2.250 & 0.893 & 0.064 & 0.959 & 2.748 & 0.932 & 0.066 & 0.953 & 1.150 & 0.424 & 0.052 & 0.957 \\
			Marigold-DC             & 2.155 & 0.875 & 0.057 & 0.968 & 2.546 & 0.981 & 0.062 & 0.963 & 1.078 & 0.411 & 0.042 & 0.972 \\\midrule
			LDCM (Ours)                      & \textbf{1.416} & \textbf{0.332} & \textbf{0.018} & \textbf{0.991} & \textbf{1.603} & \textbf{0.393} & \textbf{0.020} & \textbf{0.990} & \textbf{0.762} & \textbf{0.191} & \textbf{0.020} & \textbf{0.987} \\
			\specialrule{1.5pt}{0.8ex}{0.8ex}
		\end{tabular}%
	}
	\label{tab:depth_diffusion}
\end{table}

\begin{table}[h]
	\caption{Ablation study on the training data. We report the relative error (REL) for depth completion.}
	\centerfirst
	\resizebox{0.5\columnwidth}{!}{%
		\begin{tabular}{cccccc}
			\specialrule{1.5pt}{0.8ex}{0.8ex}
			\multirow{2}{*}{Configuration} & \multicolumn{5}{c}{Depth Completion ($\text{REL}\downarrow$)}\\\cmidrule{2-6}
			& KITTI & iBims-1 & DIODE & ETH3D & Average\\\midrule
			w/ more real data & 0.020 & \textbf{0.017} & 0.035 & \textbf{0.018} & \textbf{0.022} \\\midrule
			Ours &  \textbf{0.019} &   0.018   &     \textbf{0.033}    &   0.019     &     \textbf{0.022 } \\
			\specialrule{1.5pt}{0.8ex}{0.8ex}
		\end{tabular}%
	}
	\label{tab:training_data}
\end{table}
\begin{figure}[h]
	\includegraphics[width=\linewidth]{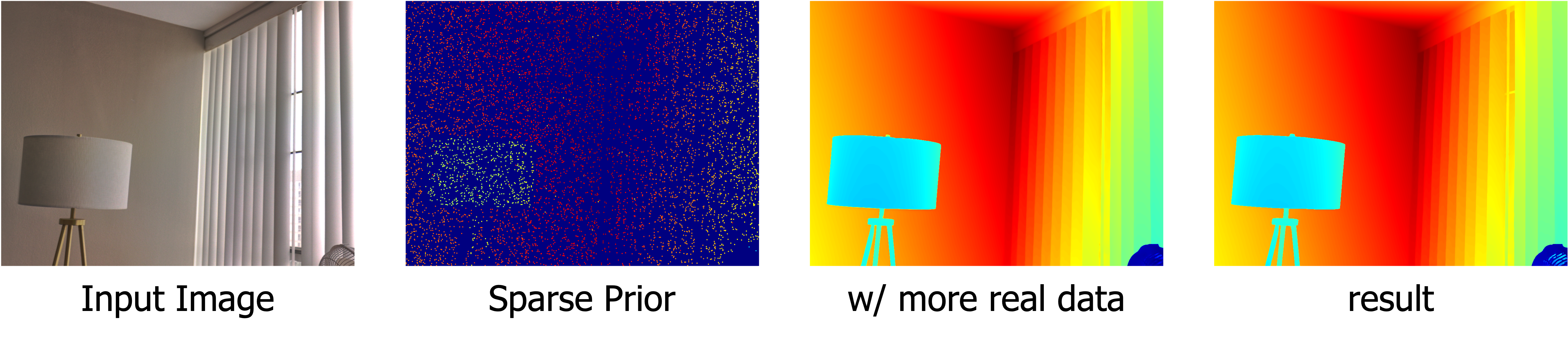}
	\caption{Qualitative comparison between the results from models using different training datasets.}
	\label{fig:training_data}
\end{figure}

\section{Ablation on the Training Data}
\textbf{Training Data.} We perform an ablation study on the training data used to train the LDCM. In addition to the original datasets, we introduce an extra dataset: DrivingStereo~\cite{drivingstereo}. The quantitative results are presented in Table~\ref{tab:training_data}. As shown, the inclusion of this additional data does not significantly affect metric performance. However, as illustrated in Fig.~\ref{fig:training_data}, incorporating more real-world data leads to visually less sharp predictions, likely due to imperfect supervision signals in the added dataset.

\section{Applying Poisson-based Alignment Strategy to Monocular Estimators}
In Table~\ref{tab:poisson_application}, we apply the Poisson alignment strategy to relative geometry estimators to obtain dense depth maps. As shown in the table, this strategy effectively improves the metric accuracy of these approaches, demonstrating its effectiveness. Moreover, our LDCM maintains state-of-the-art performance.

\begin{table}[t]
	\caption{\textbf{Quantitative comparison after applying Poisson-based alignment to relative geometry estimators.} The \colorbox{best}{best} and \colorbox{best2}{second-best} results are highlighted.}
	\resizebox{\columnwidth}{!}{%
		\begin{tabular}{cccccccccccccccc}
			\specialrule{1.5pt}{0.8ex}{0.8ex}
			\multirow{2}{*}{Method} & \multicolumn{5}{c}{KITTI}                & \multicolumn{5}{c}{iBims-1}                 & \multicolumn{5}{c}{DIODE Indoor}              \\\cmidrule(l){2-6} \cmidrule(l){7-11} \cmidrule(l){12-16} 
			& RMSE$\downarrow$ & MAE$\downarrow$   & REL$\downarrow$   & $\delta_1\uparrow$    & Rk.$\downarrow$ & RMSE$\downarrow$ & MAE$\downarrow$   & REL$\downarrow$   & $\delta_1\uparrow$  & Rk.$\downarrow$ & RMSE$\downarrow$ & MAE$\downarrow$   & REL$\downarrow$   & $\delta_1\uparrow$  & Rk.$\downarrow$ \\\midrule
			DepthAnythingV2$\dagger$ & 4.007 & 1.890 & 0.092 & 0.916 & 5.318 & 0.349 & 0.179 & 0.043 & 0.975 & 5.614 & 0.386 & 0.189 & 0.045 & \colorbox{best2}{0.976} & 4.909 \\
			DepthAnythingV2 w/ Poisson & 2.448 & 0.953 & 0.051 & \colorbox{best2}{0.959} & 2.955 & 0.231 & 0.098 & 0.027 & 0.976 & 3.136 & 0.195 & 0.091 & \colorbox{best2}{0.026} & 0.967 & 2.795 \\
			VGGT$\dagger$ & 4.219 & 2.518 & 0.158 & 0.783 & 6.955 & 0.348 & 0.194 & 0.053 & 0.957 & 6.727 & 0.425 & 0.294 & 0.096 & 0.920 & 6.750 \\
			VGGT w/ Poisson & 2.627 & 1.112 & 0.065 & 0.937 & 4.205 & 0.241 & 0.104 & 0.028 & 0.975 & 4.318 & 0.217 & 0.111 & 0.037 & 0.957 & 4.341 \\
			MoGe V1$\dagger$ & 3.050 & 1.821 & 0.125 & 0.887 & 5.341 & 0.238 & 0.120 & 0.035 & \colorbox{best2}{0.981} & 4.159 & 0.272 & 0.175 & 0.064 & 0.950 & 5.250 \\
			MoGe V1 w/ Poisson & \colorbox{best2}{2.179} & \colorbox{best2}{0.865} & \colorbox{best2}{0.050} & \colorbox{best2}{0.959} & \colorbox{best2}{2.136} & \colorbox{best2}{0.214} & \colorbox{best2}{0.089} & \colorbox{best2}{0.025} & 0.977 & \colorbox{best2}{2.409} & \colorbox{best2}{0.177} & \colorbox{best2}{0.085} & 0.028 & 0.965 & \colorbox{best2}{2.614} \\\midrule
			LDCM (Ours) & \colorbox{best}{\textbf{1.911}} & \colorbox{best}{\textbf{0.537}} & \colorbox{best}{\textbf{0.026}} & \colorbox{best}{\textbf{0.983}} & \colorbox{best}{\textbf{1.023}} & \colorbox{best}{\textbf{0.161}} & \colorbox{best}{\textbf{0.044}} & \colorbox{best}{\textbf{0.012}} & \colorbox{best}{\textbf{0.991}} & \colorbox{best}{\textbf{1.227}} & \colorbox{best}{\textbf{0.084}} & \colorbox{best}{\textbf{0.025}} & \colorbox{best}{\textbf{0.008}} & \colorbox{best}{\textbf{0.993}} & \colorbox{best}{\textbf{1.000}} \\\specialrule{1pt}{0.4ex}{0.4ex}
			\multirow{2}{*}{Method} & \multicolumn{5}{c}{DIODE Outdoor}        & \multicolumn{5}{c}{ETH3D}                  & \multicolumn{5}{c}{Average}              \\\cmidrule(l){2-6} \cmidrule(l){7-11} \cmidrule(l){12-16} 
			& RMSE$\downarrow$ & MAE$\downarrow$ & REL$\downarrow$   & $\delta_1\uparrow$ & Rk.$\downarrow$ & RMSE$\downarrow$ & MAE$\downarrow$   & REL$\downarrow$ & $\delta_1\uparrow$ & Rk.$\downarrow$ & RMSE$\downarrow$ & MAE$\downarrow$ & REL$\downarrow$ & $\delta_1\uparrow$ & Rk.$\downarrow$ \\\midrule
			DepthAnythingV2$\dagger$ & 5.940 & 2.777 & 0.124 & 0.869 & 5.114 & 2.091 & 0.424 & 0.049 & 0.979 & 5.864 & 2.555 & 1.092 & 0.071 & 0.943 & 5.364 \\
			DepthAnythingV2 w/ Poisson & 3.285 & 1.182 & 0.064 & 0.941 & 3.455 & 0.662 & 0.168 & 0.025 & 0.983 & 3.966 & 1.364 & 0.498 & 0.039 & 0.965 & 3.261 \\
			VGGT$\dagger$ & 4.898 & 2.893 & 0.237 & 0.772 & 5.705 & 0.540 & 0.317 & 0.060 & 0.949 & 5.818 & 2.086 & 1.243 & 0.121 & 0.876 & 6.391 \\
			VGGT w/ Poisson & 2.910 & 1.262 & 0.081 & 0.917 & 3.750 & 0.339 & 0.140 & 0.024 & 0.980 & 3.262 & 1.267 & 0.546 & 0.047 & 0.953 & 3.975 \\
			MoGe V1$\dagger$ & 10.576 & 8.340 & 0.406 & 0.599 & 6.932 & 1.651 & 0.550 & 0.082 & 0.943 & 5.455 & 3.157 & 2.201 & 0.142 & 0.872 & 5.427 \\
			MoGe V1 w/ Poisson & \colorbox{best2}{2.340} & \colorbox{best2}{0.910} & \colorbox{best2}{0.053} & \colorbox{best2}{0.950} & \colorbox{best2}{2.000} & \colorbox{best2}{0.319} & \colorbox{best2}{0.118} & \colorbox{best2}{0.021} & \colorbox{best2}{0.986} & \colorbox{best2}{2.262} & \colorbox{best2}{1.046} & \colorbox{best2}{0.413} & \colorbox{best2}{0.035} & \colorbox{best2}{0.967} & \colorbox{best2}{2.284} \\\midrule
			LDCM (Ours) & \colorbox{best}{\textbf{1.969}} & \colorbox{best}{\textbf{0.529}} & \colorbox{best}{\textbf{0.031}} & \colorbox{best}{\textbf{0.970}} & \colorbox{best}{\textbf{1.000}} & \colorbox{best}{\textbf{0.187}} & \colorbox{best}{\textbf{0.048}} & \colorbox{best}{\textbf{0.008}} & \colorbox{best}{\textbf{0.997}} & \colorbox{best}{\textbf{1.000}} & \colorbox{best}{\textbf{0.862}} & \colorbox{best}{\textbf{0.237}} & \colorbox{best}{\textbf{0.017}} & \colorbox{best}{\textbf{0.987}} & \colorbox{best}{\textbf{1.050}} \\
			\specialrule{1.5pt}{0.8ex}{0.8ex}
		\end{tabular}%
	}
	\label{tab:poisson_application}
\end{table}

\section{Inference Time}
We report the per-stage inference times of our method, measured at a resolution of 480×640 on an NVIDIA L20 GPU. Our pipeline comprises four stages: Depth Anything Small (0.006 s), global alignment (0.006 s), Poisson-based alignment (0.020 s), and the subsequent refinement model (0.040 s), resulting in a total runtime of 0.072 s. For comparison, LWLR runs in 0.010 s under the same conditions. A detailed comparison with the inference times of several competing methods is provided in Table~\ref{tab:inferencetime}.

\begin{table}[h]
	\caption{Inference time (in seconds) of different methods at $480 \times 640$ resolution on an NVIDIA L20 GPU, with all inference performed in FP32 precision.}
	\resizebox{\columnwidth}{!}{%
		\begin{tabular}{cccccccc}
			\specialrule{1.5pt}{0.8ex}{0.8ex}
			Method & OMNI-DC & PriorDA & DepthPro & VGGT & MoGe V2 & DepthAnythingV2 & LDCM (Ours) \\\cmidrule{2-8}
			Inference Time (s) & 0.128 & 0.064 & 0.554 & 0.196 & 0.220 & 0.019 & 0.072\\
			\specialrule{1.5pt}{0.8ex}{0.8ex} 
		\end{tabular}%
	}
	\label{tab:inferencetime}
\end{table}

\begin{figure}[t]
	\includegraphics[width=\linewidth]{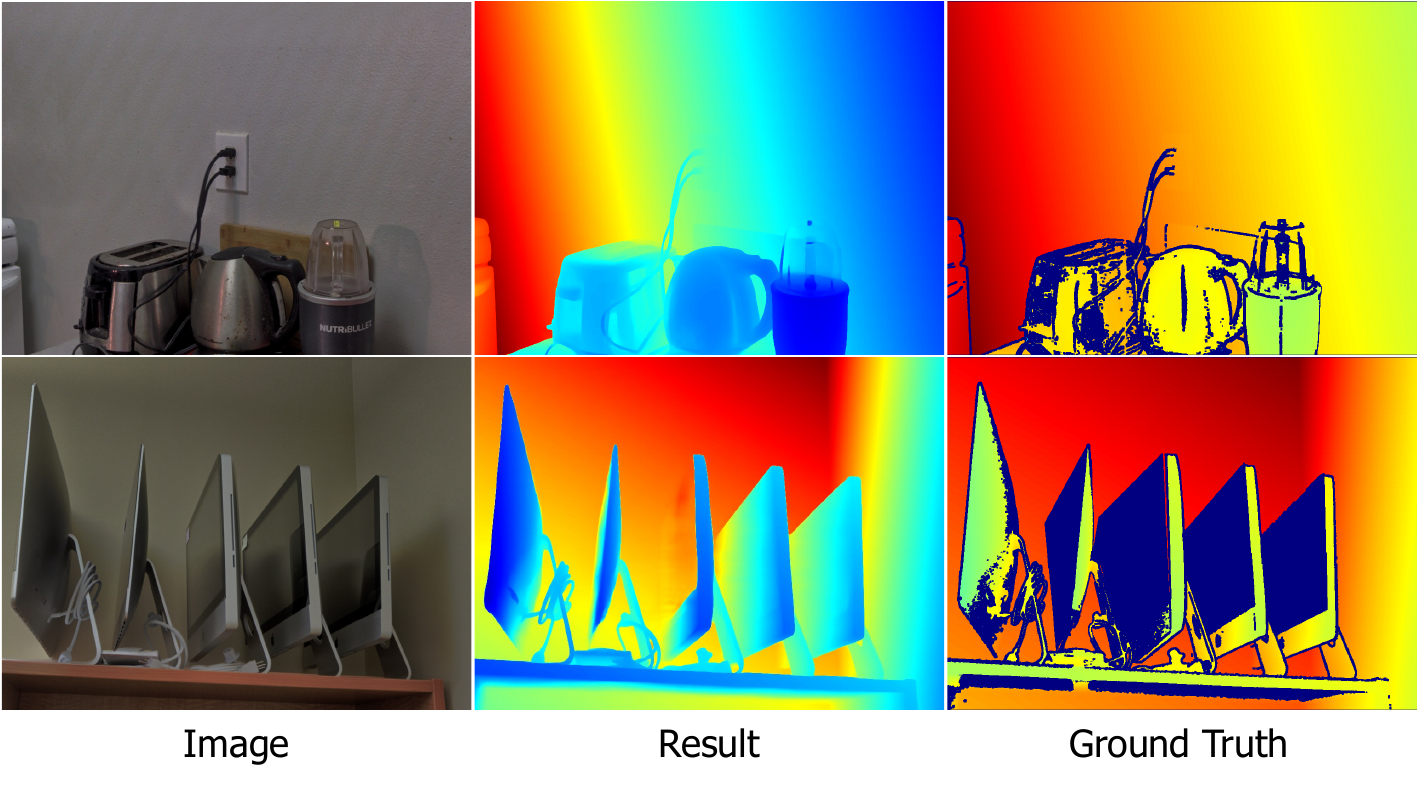}
	\caption{Examples of two failure cases.}
	\label{fig:failure}
\end{figure}

\section{More Qualitative Results}
\begin{figure}[t]
	\includegraphics[width=\linewidth]{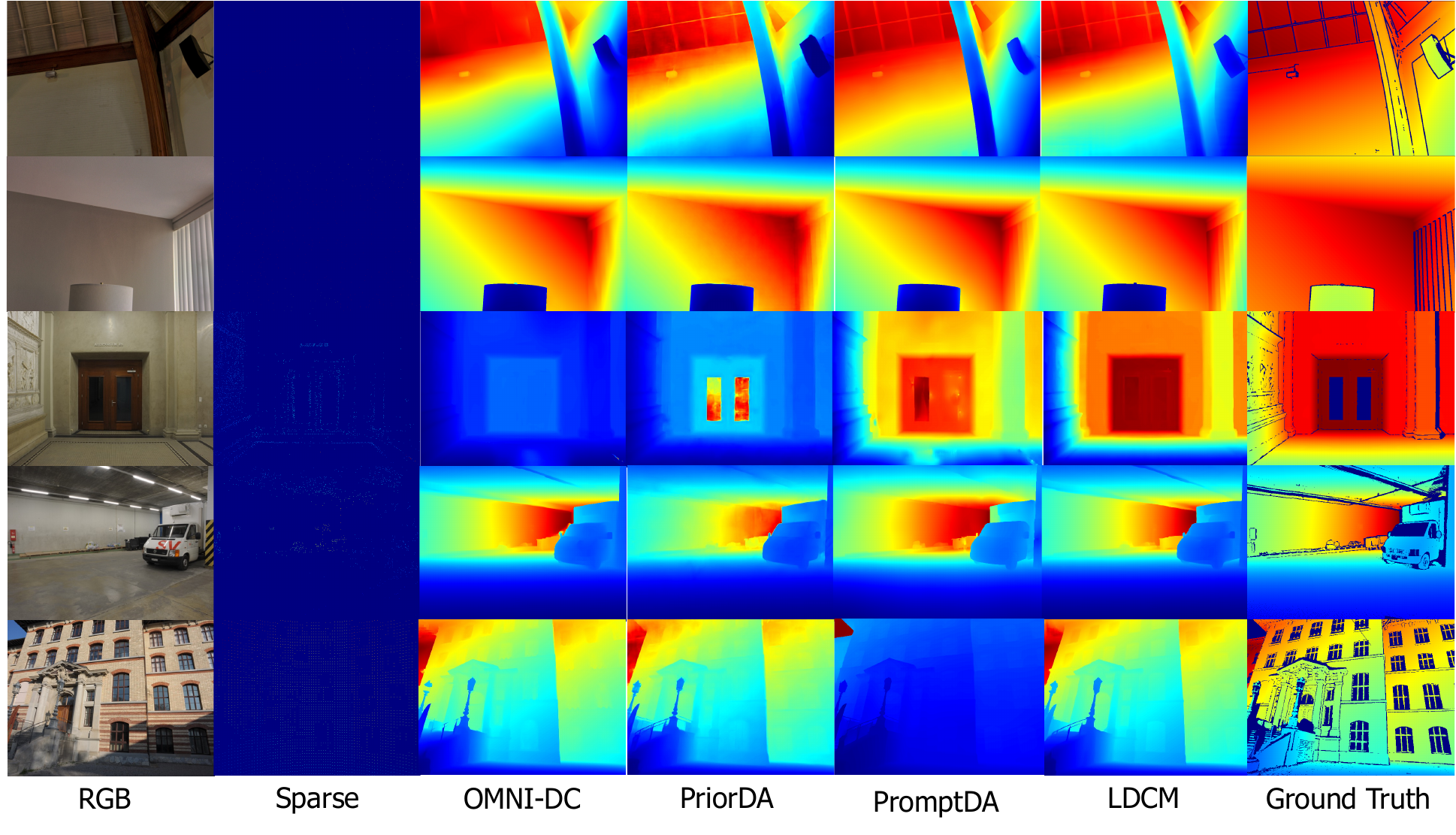}
	\caption{Visualization comparison with state-of-the-art methods.}
	\label{fig:qualiative}
\end{figure}

\begin{figure}[t]
	\includegraphics[width=\linewidth]{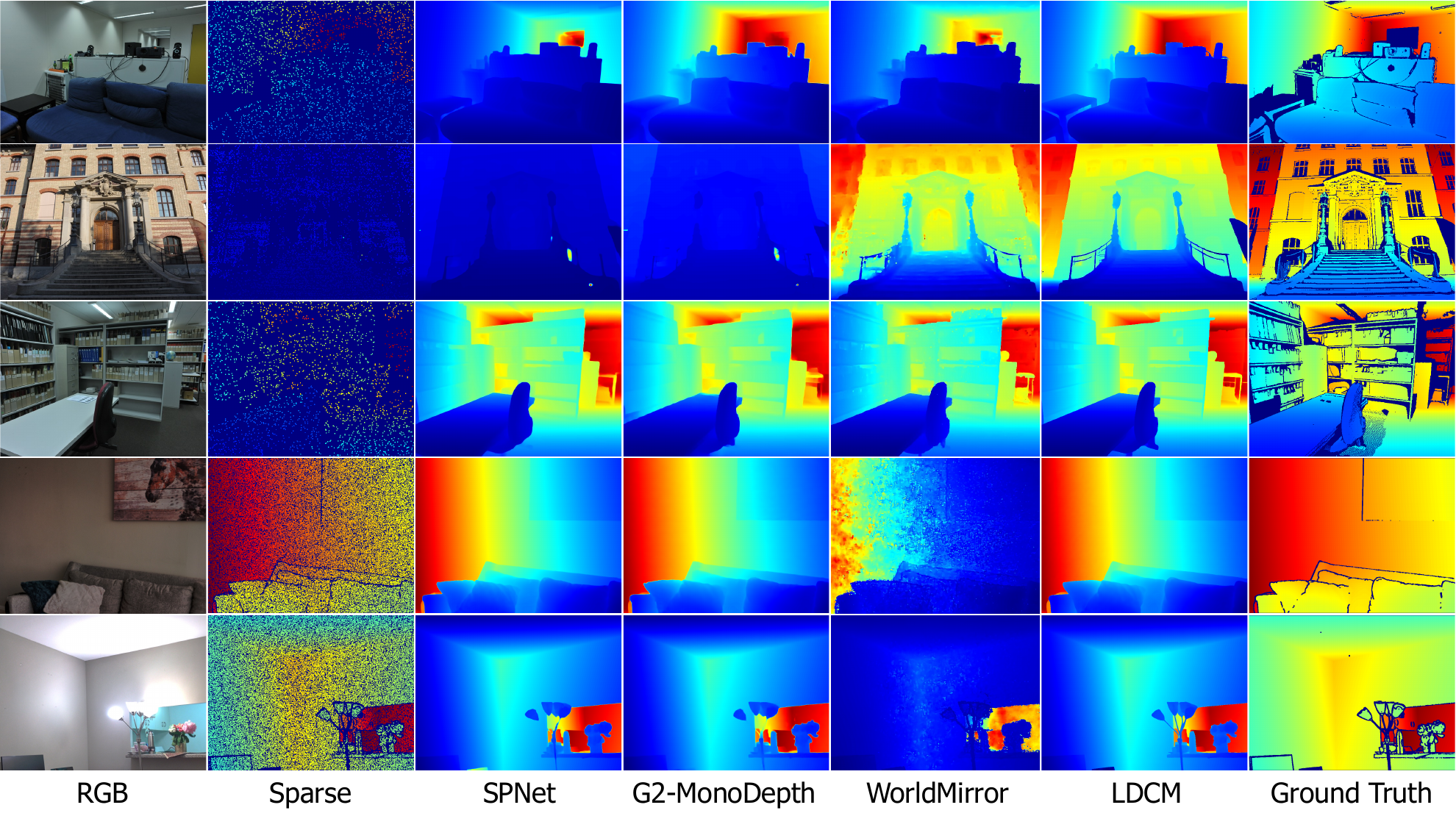}
	\caption{Visualization comparison with state-of-the-art methods.}
	\label{fig:qualiative2}
\end{figure}

\begin{figure}[t]
	\includegraphics[width=\linewidth]{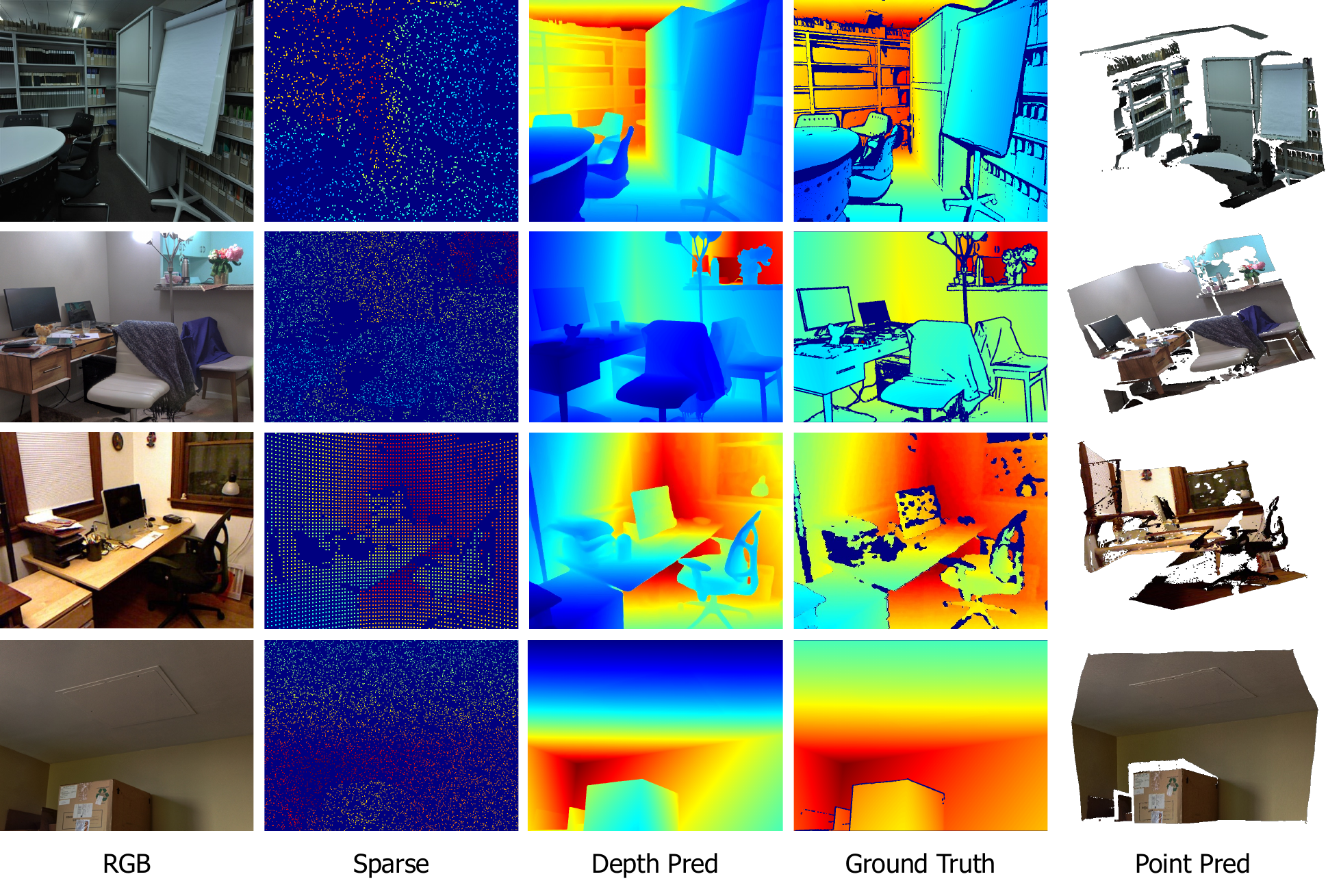}
	\caption{More visualization results for depth map and point map.}
	\label{fig:qualiative_point}
\end{figure}

Fig.~\ref{fig:qualiative} and Fig.~\ref{fig:qualiative2} present a qualitative comparison between LDCM and state-of-the-art methods. 
Notably, LDCM produces sharper geometric structures and more accurate depth distributions, particularly in regions with complex geometry or extreme sparsity. The predictions from LDCM exhibit significantly clearer boundaries and finer details, demonstrating the effectiveness of our coarse-to-fine framework and structural prior integration. In Fig.~\ref{fig:qualiative_point}, we provide more visualization results for depth map and point map.

\section{Noise Analysis} 
\begin{figure}[h]
	\includegraphics[width=\linewidth]{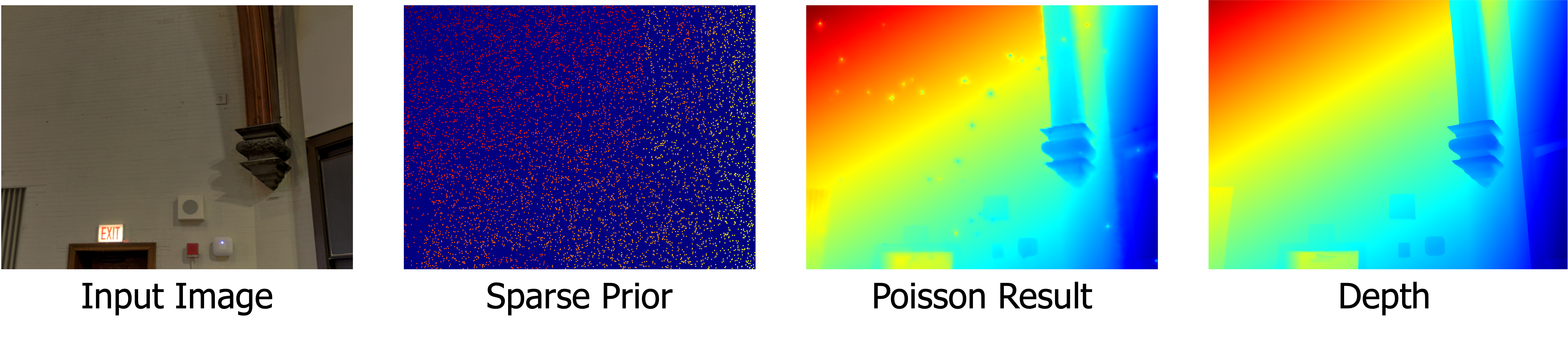}
	\caption{An example with noisy input.}
	\label{fig:noise}
\end{figure}

Fig.~\ref{fig:noise} presents an example with noisy input. When the sparse prior contains noise, the Poisson alignment strategy is adversely affected. However, the subsequent network effectively mitigates this issue and produces a high-quality result.

\section{Limitation and Future Work}
Although LDCM achieves superior performance, accurately reconstructing transparent objects and reflective surfaces remains challenging, as illustrated by two failure cases in Fig.~\ref{fig:failure}. This limitation stems from the lack of large-scale datasets containing such materials, which are difficult to capture and annotate. In the future, we plan to investigate synthetic data simulation to augment training and improve robustness on these challenging scenarios. Additionally, while monocular video reconstruction is a promising application, achieving temporal consistency poses substantial challenges. Extending LDCM to process video sequences for consistent 3D geometry estimation over time is an important direction for future exploration.

\section{Statement on the Use of LLMs} 
Large language models (LLMs) were used only for linguistic refinement, such as improving grammar and phrasing. They played no role in shaping research concepts, designing experiments, or interpreting data. The authors authored all content, verified its accuracy and originality, and assume full responsibility for the manuscript.

\end{document}